\documentclass[preprint,review,12pt]{elsarticle}
\usepackage{graphicx}

\usepackage{amsmath,amsfonts,bm}

\def\eqref#1{equation~\ref{#1}}

\def\1{\bm{1}}

\DeclareMathAlphabet{\mathsfit}{\encodingdefault}{\sfdefault}{m}{sl}
\SetMathAlphabet{\mathsfit}{bold}{\encodingdefault}{\sfdefault}{bx}{n}

\usepackage{tikz}
\usepackage{comment}
\usepackage{amsmath,amssymb} 
\usepackage{color}

\usepackage[accsupp]{axessibility}  

\usepackage{soul}
\usepackage[hidelinks]{hyperref}
\usepackage[small]{caption}
\usepackage{booktabs}
\usepackage{algorithm}  
\usepackage[noend]{algpseudocode}
\usepackage{amsfonts}
\usepackage{bm}
\usepackage{subcaption}

\usepackage{multirow}
\usepackage{rotating}

\journal{Neurocomputing}

\makeatletter
\DeclareRobustCommand\onedot{\futurelet\@let@token\@onedot}
\def\@onedot{\ifx\@let@token.\else.\null\fi\xspace}

\def\eg{\emph{e.g.}} 
\def\ie{\emph{i.e.}}

\makeatother

\begin{document}

\begin{frontmatter}

\title{Label-Efficient Object Detection via Region Proposal Network Pre-Training}

\author[1]{Nanqing Dong\corref{cor1}\fnref{fnt1}}
\cortext[cor1]{Corresponding authors}
\ead{dongnanqing@pjlab.org.cn}
\fntext[fnt1]{The first two authors contributed equally.}
\author[2]{Linus Ericsson\fnref{fnt1}}
\author[3]{Yongxin Yang}
\author[4]{Ales Leonardis}
\author[2]{Steven McDonagh\corref{cor1}}
\ead{s.mcdonagh@ed.ac.uk}
\affiliation[1]{organization={Shanghai Artificial Intelligence Laboratory},
            city={Shanghai},
            postcode={200232}, 
            country={China}}
\affiliation[2]{organization={Institute for Imaging, Data and Communications (IDCOM), School of Engineering, University of Edinburgh},
            city={Edinburgh},
            postcode={EH9 3FG},
            country={UK}}
\affiliation[3]{organization={School of Electronic Engineering and Computer Science, Queen Mary University of London},
            city={London},
            postcode={E1 4NS},
            country={UK}}
\affiliation[4]{organization={School of Computer Science, University of Birmingham},
            city={Birmingham},
            postcode={B15 2TT},
            country={UK}}

\begin{abstract}
Self-supervised pre-training, based on the pretext task of instance discrimination, has fueled the recent advance in label-efficient object detection. However, existing studies focus on pre-training only a feature extractor network to learn transferable representations for downstream detection tasks. This leads to the necessity of training multiple detection-specific modules from scratch in the fine-tuning phase. We argue that the region proposal network (RPN), a common detection-specific module, can additionally be pre-trained towards reducing the localization error of multi-stage detectors.
In this work, we propose a simple pretext task that provides an effective pre-training for the RPN, towards efficiently improving downstream object detection performance. We evaluate the efficacy of our approach on benchmark object detection tasks and additional downstream tasks, including instance segmentation and few-shot detection. In comparison with multi-stage detectors without RPN pre-training, our approach is able to consistently improve downstream task performance, with largest gains found in label-scarce settings.

\begin{keyword}
Self-supervised learning \sep Object detection
\end{keyword}

\end{abstract}

\end{frontmatter}


\section{Introduction}
\label{sec:intro}

Image-level representation learning based on instance discrimination has proven highly effective as a general model for transfer learning, successfully transferring to diverse downstream tasks~\cite{he2020momentum,grill2020bootstrap,caron2021emerging,ericsson2021well}. Such generality, however, comes at a cost; specificity is sacrificed in cases where a target downstream task is known apriori, \ie~where \emph{task misalignment} exists~\cite{zhang2016colorful,dong2021self}. In particular, it has been noted that image-level learned representations are inadequate for object-level downstream tasks such as object detection and instance segmentation~\cite{wang2021dense,yang2021instance,henaff2021efficient}. 

Object detection poses a fundamental and challenging computer vision problem. Solving detection tasks successfully requires combination of (1) finding object locations, and (2) classifying located objects. Popular contemporary supervised object detection commonly tackles this by employing multi-stage architectures that suggest object proposals (locations) which a sequential network must then recognise. In contrast, most existing self-supervised pre-training for object detection tasks focus solely on solving component (2); through approximation of a classification loss~\cite{xie2021detco,xie2021propagate,chen2021multisiam,wei2021aligning}. This leads to situations where classification-related architectural components receive gradient updates from both unlabeled (pre-training) and labeled (fine-tuning) data, while localization-related components receive gradient updates only from labeled data. We conjecture that such training setups are sub-optimal for overall performance, especially in cases offering only limited labeled data.
\begin{figure}[t]
    \centering
    \includegraphics[width=1.0\columnwidth]{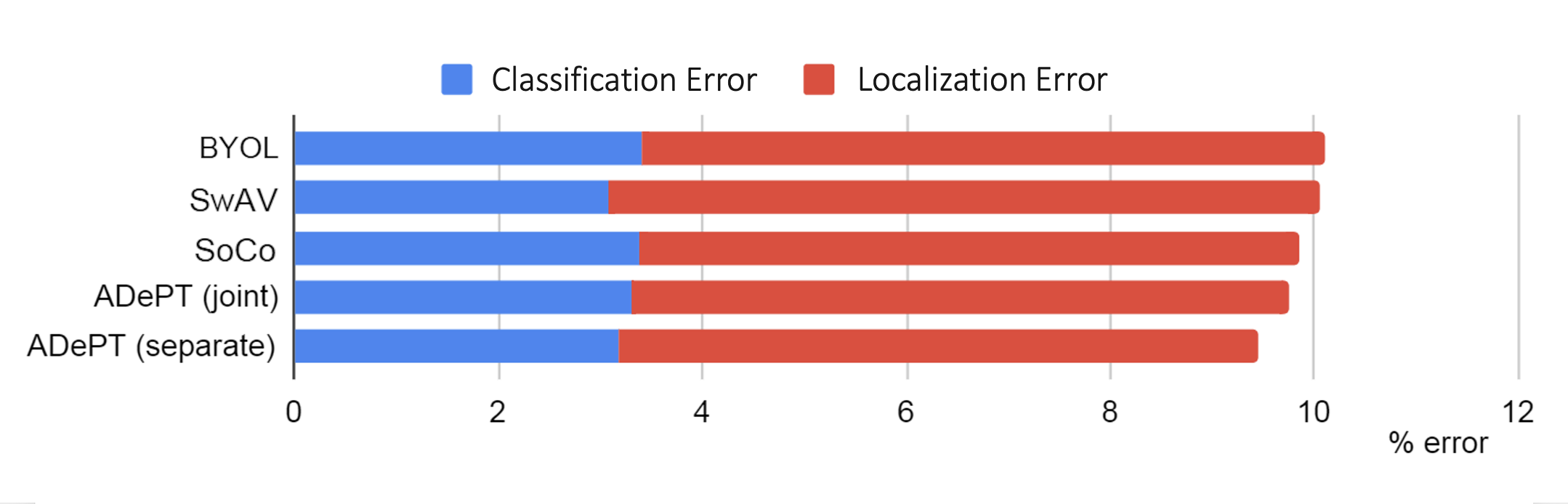}
    \caption{Localization errors contribute significantly to overall detection error rates. The effect is observed by evaluating recent SSL approaches BYOL~\cite{grill2020bootstrap}, SwAV~\cite{caron2020unsupervised}, SoCo~\cite{wei2021aligning}. 
    Our ADePT method reduces the dominant localization error term. See text for further details.
    }
    \label{fig:cls_and_loc_errors}
\end{figure}

\begin{figure}[th]
    \centering
    \begin{subfigure}[t]{0.24\textwidth}
        \centering
        \includegraphics[width=1\textwidth]{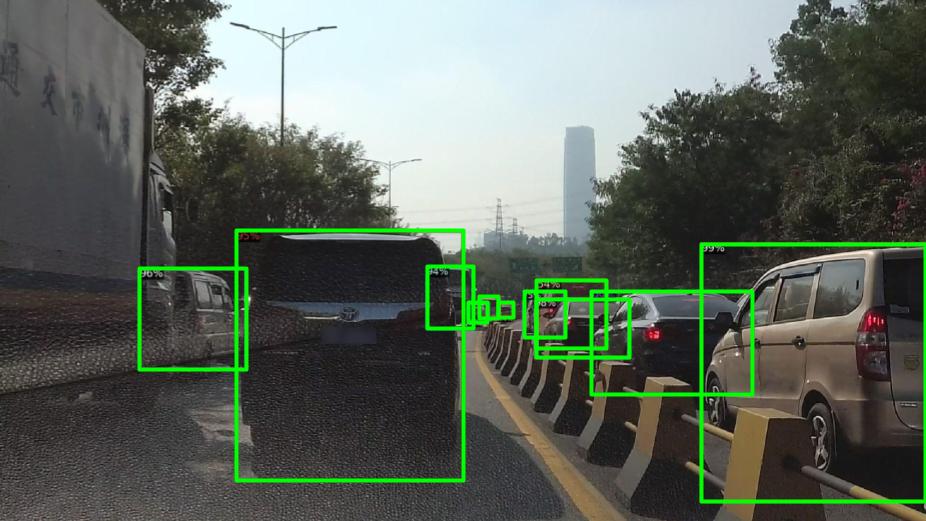}
        \includegraphics[width=1\textwidth]{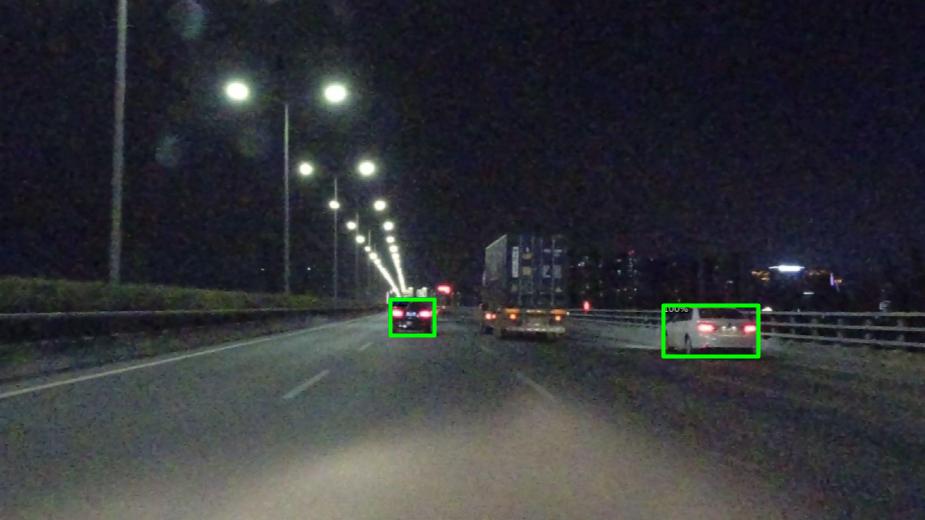}
        \caption{BYOL}
    \end{subfigure}
    \begin{subfigure}[t]{0.24\textwidth}
        \centering
        \includegraphics[width=1\textwidth]{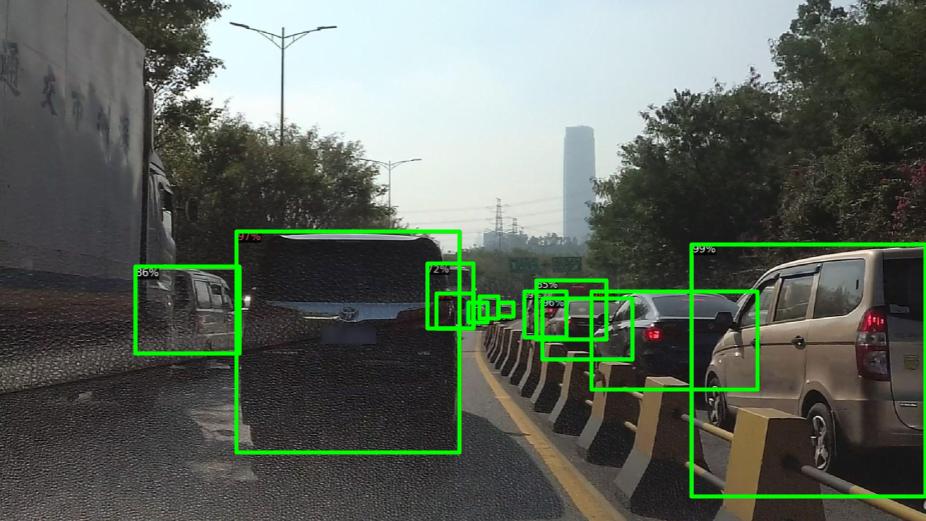}
        \includegraphics[width=1\textwidth]{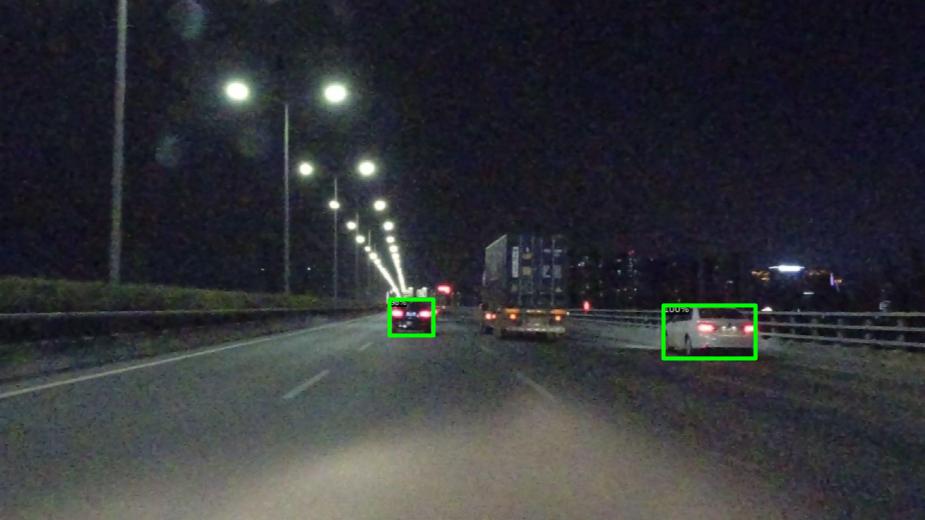}
        \caption{SwAV}
    \end{subfigure}
    \begin{subfigure}[t]{0.24\textwidth}
        \centering
        \includegraphics[width=1\textwidth]{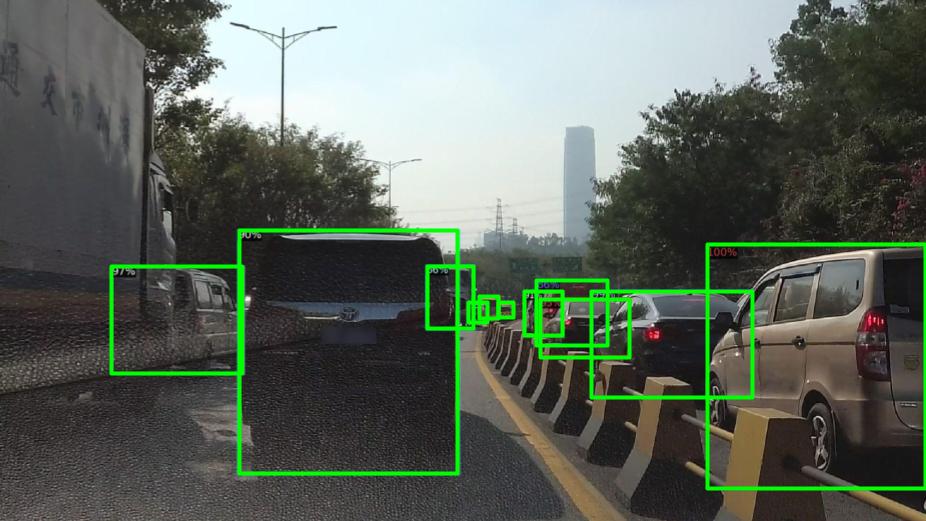}
        \includegraphics[width=1\textwidth]{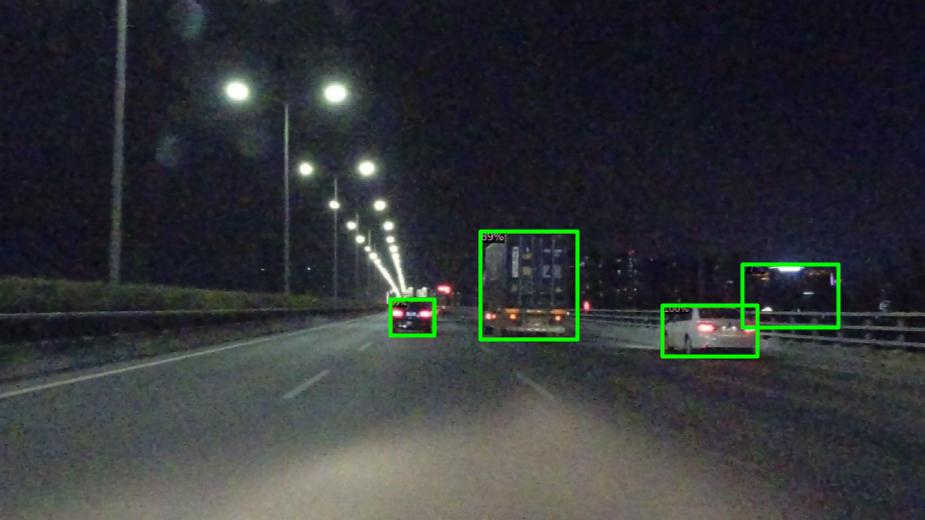}
        \caption{SoCo}
    \end{subfigure}
    \begin{subfigure}[t]{0.24\textwidth}
        \centering
        \includegraphics[width=1\textwidth]{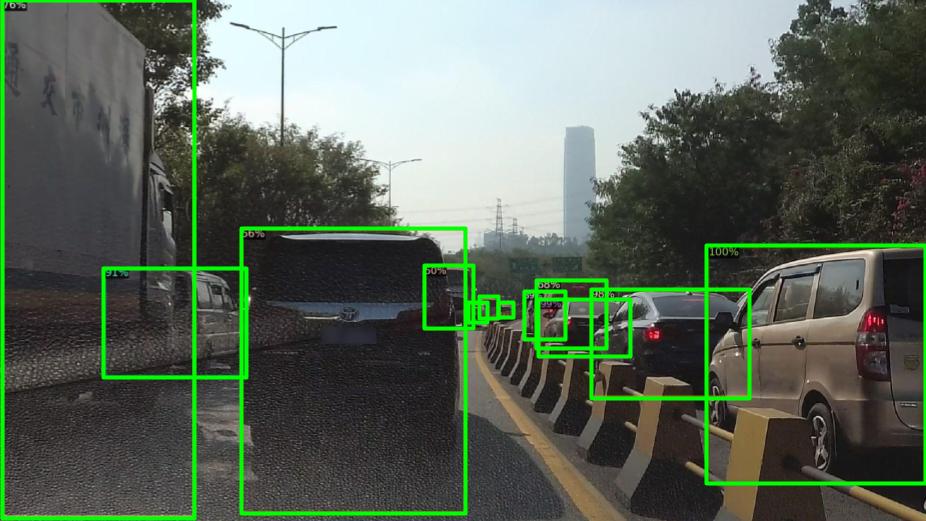}
        \includegraphics[width=1\textwidth]{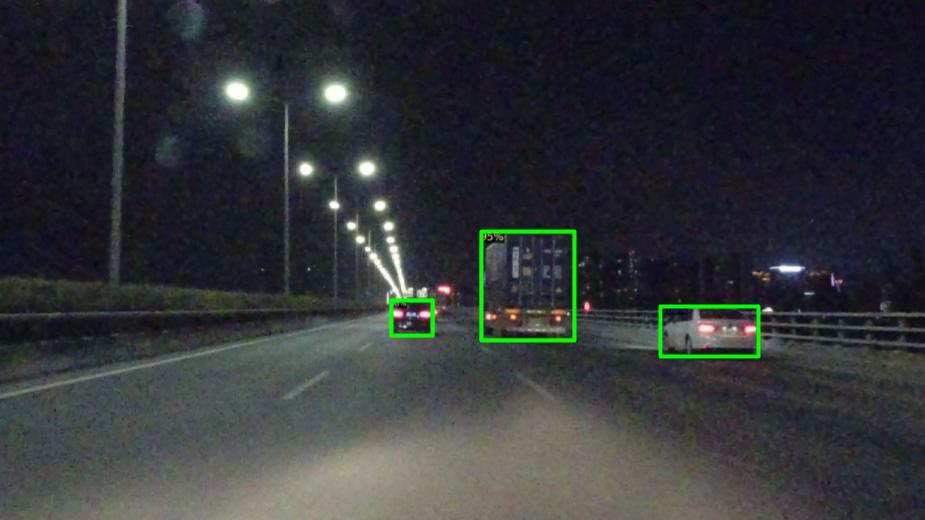}
        \caption{ADePT (ours)}
    \end{subfigure}
    \caption{Qualitative comparison using SODA10M~\cite{han2021soda10m}. Top row: ADePT is the {only} method able to detect the partially occluded truck (image left). Bottom row: dark background and low illumination make successful detection challenging. BYOL~\cite{grill2020bootstrap} and SwAV~\cite{caron2020unsupervised} fail to detect all vehicles. SoCo~\cite{wei2021aligning} and ADePT successfully capture {all} three, yet SoCo hallucinates an object on the right hand side. \textbf{Best viewed with digital zoom}.}
    \label{fig:qualitative_soda}
\end{figure}

A further object detection challenge, largely agnostic to the employed learning strategy, can be attributed to the opaqueness of standard mAP metrics: disambiguating different \emph{types} of detection error can be difficult. We find that inspection of particular error types, in relation to overall system performance, allows for minimization of confounding variables and also motivates us to focus our efforts on sources of detection error where most meaningful improvements can be made. Existing analysis of detection error sources~\cite{bolya2020tide} highlights that self-supervised methods suffer from a higher prevalence of \emph{localization} errors in comparison with \emph{classification} errors, a result we corroborate in this work, by pre-training various models on ImageNet~\cite{deng2009imagenet} and fine-tuning on MS COCO using Mask R-CNN~\cite{he2017mask} \mbox{(see Fig.~\ref{fig:cls_and_loc_errors})}, full results are found in Table~\ref{tab:errors}. This further motivates the idea of improving upon multi-stage detector components where causal relationships with object localization performance can be identified.

A promising recent design principle involves encouraging alignment between the pretext task and the downstream task. SoCo~\cite{wei2021aligning} achieves state-of-the-art performance by pre-training model components consisting of feature extractor and detection head components. Our insight into the importance of localization-based errors motivates us to extend these pre-training concepts, to further evaluate whether model components based on region proposal networks (RPN)~\cite{ren2015faster} can provide meaningful performance improvement, particularly in situations with few labelled instances. Directly addressing the aforementioned localization error issues has, until now, remained largely missing from self-supervised approaches.

We name our approach \textbf{A}ligned \textbf{De}tection \textbf{P}re-\textbf{T}raining (ADePT). In contrast to supervised detection training setups, ADePT does not require: ($a$) ground-truth class labels; classification is approximated by treating cropped regions as individual classes and solving an instance discrimination task, and further does not require: ($b$) ground-truth bounding boxes; we will show that location labels can be successfully approximated via unsupervised region proposals, generated by heuristic algorithms. We instantiate the latter idea in this work using selective search~\cite{uijlings2013selective}.

Through evaluation of our ideas, we aim to answer the following research questions: 
\textbf{Q1.}~Can the inclusion of RPN module pre-training reduce (commonly high) localization-based detection error terms?
\textbf{Q2.}~Does principled alignment of pretext and downstream tasks improve overall performance? 
\textbf{Q3.}~Does such alignment enable and benefit more label-efficient scenarios? 

We evaluate ADePT on three benchmark detection tasks: MS COCO~\cite{lin2014microsoft}, SODA10M~\cite{han2021soda10m}, and PASCAL VOC~\cite{everingham2010pascal}. We find that ADePT consistently provides stronger downstream detection performance over several seminal self-supervised pre-training baselines, especially in domains containing only limited labeled data (\eg~when only $1\%$ and $10\%$ of labels are available). A qualitative comparison is presented in Fig.~\ref{fig:qualitative_soda}. Additionally, we observe boosts in few-shot detection, highlighting a further benefit of comprehensively pre-trained architectures.

Our main contributions can be summarized as:
\begin{itemize}
    \item We explore, for the first time, pre-training of RPN components in two-stage detectors.
    \item Empirical analysis of \mbox{\emph{localization}} error terms provide new insights for self-supervised object detection research considerations. 
    \item Experimental work across several benchmarks and settings comprehensively evidences the benefits of RPN pre-training in terms of transferable representation. 
\end{itemize}

\section{Related Work} 
\label{sec:related_work}

\subsection{Multi-Stage Object Detection} 
Recent success in supervised object detection has been largely driven by multi-stage architectures that combine region proposal methods with region-based convolutional neural networks (R-CNNs), in order to produce classification and localization predictions~\cite{girshick2014rich,girshick2015fast,ren2015faster,he2017mask,cai2018cascade}. 
The capstone Faster R-CNN~\cite{ren2015faster} proposes the region proposal network (RPN); a fully convolutional network that simultaneously predicts bounding boxes and objectness scores at each spatial location, thus enabling learning of high quality proposals while sharing representations with the detector head. Notable further extensions to this architecture include Mask R-CNN~\cite{he2017mask} and Cascade R-CNN~\cite{cai2018cascade}. 
In this work we draw inspiration from the aforementioned advances found in the supervised object detection literature and introduce a pretext task capable of pre-training both R-CNN and RPN multi-stage detector components, towards gaining maximal benefit from the capabilities of these architectural ideas under the \mbox{self-supervised} paradigm. 

\subsection{Self-Supervised Learning for Object Detection}
The renascence of self-supervised learning (SSL) provides an effective initialization strategy, alternative to supervised pre-training, that is capable of learning useful representations from unlabeled data. Instance discrimination-based methods are now able to efficiently pre-train feature extractor components resulting in strong downstream performance in classification tasks, competitive with supervised pre-training~\cite{henaff2020data,he2020momentum,chen2020simple,grill2020bootstrap}. Such pre-training strategies have also been shown to boost object detection performance in label-scarce settings~\cite{he2020momentum,grill2020bootstrap}. For example, in BYOL~\cite{grill2020bootstrap}, two augmented views are processed by an online branch and a target branch with the same network architecture. The parameters of the target branch are the moving average of the parameters of online branch with stop-gradient. A similarity loss is minimized between two views to optimize the online branch.
The state-of-the-art method, SoCo~\cite{wei2021aligning}, extends the strategy beyond solitary feature extractor pre-training and includes also detector head alignment, which consists of a feature pyramid network (FPN)~\cite{lin2017feature} and region of interest (RoI) head~\cite{girshick2015fast}. Object proposals (\ie~bounding boxes) are generated in a heuristic fashion using selective search~\cite{uijlings2013selective}, allowing object-level (rather than global image-level) features to be contrasted. 
Although the detector head components are pre-trainable, SoCo~\cite{wei2021aligning} disregards the RPN. Similarly, although~\cite{bai2022point,islam2023self} quantitatively outperform SoCo through the proposal of improved contrastive learning strategies, none of these studies pay attention to the pre-training of the RPN component. This omission results in a requirement to train the RPN from scratch with data that contains downstream task labels. In this work, we demonstrate that pre-training RPN components with an appropriate pretext task can further improve performance, especially with respect to localization error reduction.

\section{Method} 
\label{sec:method}

\begin{figure}
    \centering
    \includegraphics[width=\textwidth]{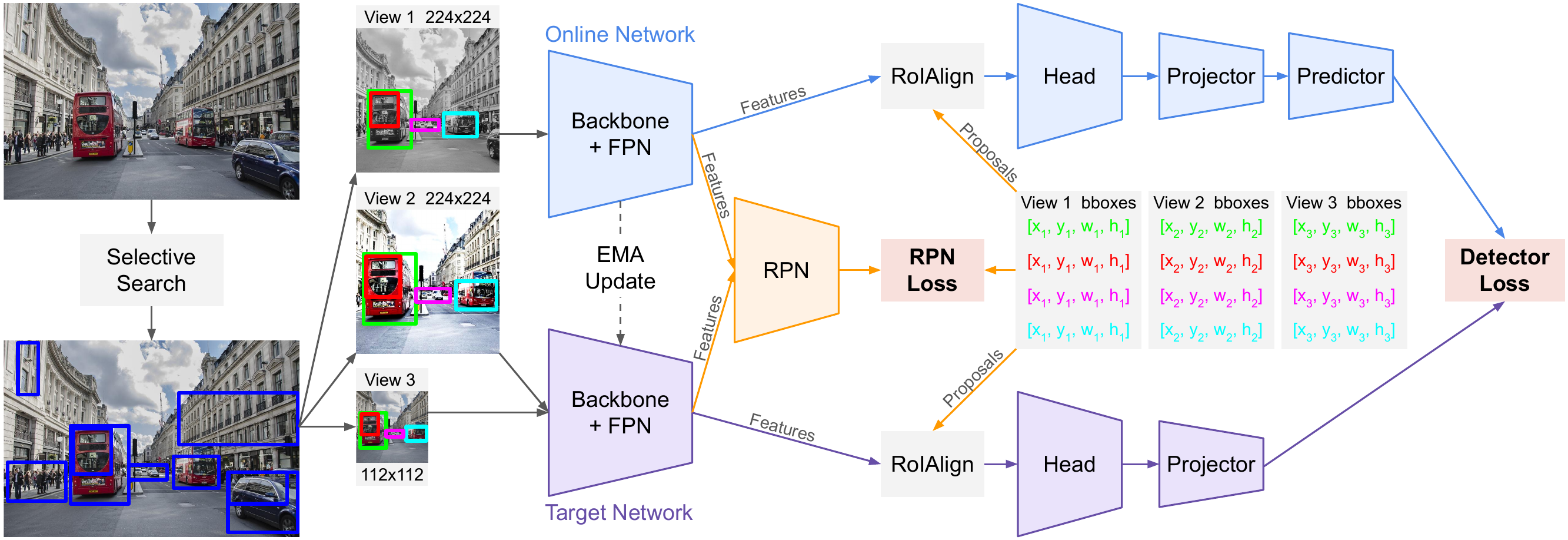}
    \caption{Diagram of ADePT. We perform self-supervised pre-training of the RPN component (Sec.~\ref{sec:method:rpn},~\ref{sec:method:ssl_rpn}). The feature extractor (backbone + FPN) and detector head are also trained in self-supervised fashion where three augmented views of the same region in the original image are passed through an online branch (blue) and a target branch (purple), to form a contrastive detector 
    loss.}
    \label{fig:diagram}
\end{figure}

We follow common two-stage object detector architectural design that consists of three main architectural components: a feature extractor, an RPN, and a detector head. The feature extractor contains a backbone (\eg~ResNet50~\cite{he2016deep}) and a feature pyramid network (FPN)~\cite{lin2017feature}. The training of the feature extractor and detector head follows a BYOL-style contrastive learning strategy~\cite{grill2020bootstrap}, where two networks with identical architectures are denoted \emph{online} and \emph{target} networks, respectively. In the training phase, coarse bounding boxes are first generated by selective search~\cite{uijlings2013selective}. The input image and generated bounding boxes are then used to generate three augmented views. The augmentation details are provided in Sec.~\ref{sec:exp:aug}. These views are passed through the feature extractor and the resulting features are fed to the RPN to generate region proposals. In this way we uniquely train the RPN to regress the selective search bounding boxes. 
The refined features are then extracted from the learned proposal regions using ``{RoIAlign}''~\cite{he2017mask}. 
Analogous to BYOL, the refined features are fed through the detector head, projector, predictor and subsequently contrasted in the detector loss which follows the SoCo formulation and is defined in Sec.~\ref{app:soco:ssl_detector}. The total loss is a combination of the RPN and detector losses, together pre-training the entire architecture. The online network receives gradient updates while the target network is maintained as the exponential moving average (EMA) of the online network. In the inference phase, only the target branch is kept and the prediction is made by feeding an input image through the target feature network, the RPN, and the target detector head sequentially. Our complete ADePT pipeline is illustrated in Fig.~\ref{fig:diagram}.

\subsection{Region Proposal Network}
\label{sec:method:rpn}
The region proposal network (RPN)~\cite{ren2015faster} component of two-stage object detectors consists of a fully convolutional network that simultaneously predicts object bounds and objectness scores at each spatial position. It works by placing a large number of \emph{anchor boxes} on a spatial grid, from which it selects those most likely to contain an object, while adjusting their position and size to more closely match the object outline. The anchors are a set of fixed size / ratio rectangles, located at each point in the feature space. The RPN feeds the features extracted from the feature extractor (denoted as ``backbone + FPN'' in Fig.~\ref{fig:diagram}) through a set of convolutional layers to predict whether or not each such anchor corresponds to an object, in addition to anchor deltas that offset shape and position. By suppressing overlapping predictions and ranking the proposals by their ``objectness'' scores, the RPN can output a set of high quality proposals. More formally, let the feature extractor be $f$, the RPN convolutional layers, $c$, the objectness layer $o$, and anchor delta layer $d$. Then given an image $x$, the objectness scores are computed as $p = o \circ c \circ f(x)$ and the anchor deltas as $t = d \circ c \circ f(x)$. The loss to train the RPN is defined as
\begin{equation}
\begin{split}
        \mathcal{L}_{\text{RPN}}(p, t) &= \frac{1}{N_{\text{cls}}} \sum_{i=1}^{N_{\text{cls}}} \mathcal{L}_{\text{cls}}(p_i, p_i^*) \\
        & + \lambda \frac{1}{N_{\text{reg}}} \sum_{i=1}^{N_{\text{reg}}} p_i^* \mathcal{L}_{\text{reg}}(t_i, t_i^*),
\end{split}
\label{eq:rpn_loss}
\end{equation}
where $p_i$ is the predicted objectness probability of anchor $i$ and $p_i^*$ is its ground-truth label. This label is $1$ if either: it has highest IoU with a ground-truth box, or it has an IoU greater than $0.7$ with any box. The objectness label is $0$ if the IoU is less than $0.3$, and any other anchors are ignored. The classification loss is a log loss over the binary objectness values. Additionally, $t_i$ are the predicted anchor deltas that adjust the position and size of the anchor. The corresponding ground-truth deltas $t_i^*$ are computed for the ground-truth box associated with the positive anchor $i$. The regression loss is a smooth $\mathcal{L}_1$ loss which only activates when $p_i^* = 1$ and is otherwise disabled. For further details, see~\cite{ren2015faster}.

\subsection{RPN Pre-Training}
\label{sec:method:ssl_rpn}
In supervised object detection, ground-truth bounding boxes, as annotated by humans, are used to train the RPN using the loss described in Sec.~\ref{sec:method:rpn}. Given our self-supervised setup, we rather mitigate a lack of ground-truth bounding boxes by alternatively considering pseudo ground-truth labels; we utilize bounding boxes generated via an unsupervised region proposal method. We here employ the selective search algorithm~\cite{uijlings2013selective} for this purpose. This algorithm uses heuristic color and texture similarity scores to create a set of bounding boxes for an input image. These bounding boxes are pre-computed before training and filtered such that only boxes that are large enough and within certain aspect ratio constraints are retained. The pretext task for pre-training the RPN is then to regress the bounding box predictions to the pre-computed bounding boxes generated by the selective search. During training, we randomly select $K$ bounding boxes for each image in a batch, and treat them as ground-truth bounding boxes towards providing a valid learning signal to update the RPN.

In the supervised detection literature, it is common to train RPN modules either \emph{jointly}, in conjunction with other model architectural components, or \emph{separately}, in a potentially alternating fashion~\cite{ren2015faster}. We experiment with analogous strategies pertaining to these options for our alternative unsupervised setting (see Sec.~\ref{sec:experiments:training}). It is worth mentioning that the aforementioned selective search strategy is only implemented in the pre-training stage. In the fine-tuning or inference stage, the proposals are generated by the RPN.

\subsection{Detector Head}
\label{app:soco:detector}
We train a Fast R-CNN detector head architecture~\cite{girshick2015fast}, using a self-supervised similarity loss. Recall that we define a feature extractor network $f_{\theta}$ and then define the Fast R-CNN layers as $g_{\theta}$. The standard output layers of the detector are replaced by network heads pertaining to a projector $p_{\theta}$ and a predictor, $q_{\theta}$. An ``offline'' encoder of the same architecture (\ie~defined as the \emph{target} network in MoCo~\cite{he2020momentum}) is maintained with parameters $\xi$. The analogous modules; $f_\xi$, $g_\xi$, $p_\xi$ are updated as an exponential moving average of the \emph{online} network parameters $\theta$. Note that the momentum encoder does not require a final predictor head, nor does it include RPN layers.

Let $b = \{b_i\}_{i=1}^ K$ be the $K$ randomly chosen bounding boxes for an image as generated by selective search. Next, given views $V_1$, $V_2$, $V_3$ and each bounding box $b_i$, we extract the object-level representation from the backbone through the \texttt{RoIAlign} method~\cite{he2017mask}
\begin{align}
    h_{i, \theta}   = g_\theta(\texttt{RoIAlign}(f_\theta(V_1), b_i)), \\
    h_{i, \xi}'  = g_\xi(\texttt{RoIAlign}(f_\xi(V_2), b_i)), \\
    h_{i, \xi}'' = g_\xi(\texttt{RoIAlign}(f_\xi(V_3), b_i)).
\end{align}
Note, in Mask R-CNN, \texttt{RoIAlign} takes the refined bounding boxes output by the RPN.
The projector and predictor heads are then used to create the latent representations of the views
\begin{align}
    v_{i, \theta} = q_\theta(p_\theta(h_i)), \quad v_{i, \xi}' = p_\xi(h_i'), \quad v_{i, \xi}'' = p_\xi(h_i'').
\end{align}
The full loss for the detector head contrasts the latent representations of the same proposal across different views by maximising their similarity.
\begin{equation}
\begin{split}
        &\mathcal{L}_{\text{sim}}(v_{\theta},  v_{\xi}',  v_{\xi}'') = \\
        &\frac{1}{K} \sum_{i=1}^K \Bigg( - \cdot \frac{2 \langle v_{i, \theta}, v_{i, \xi}' \rangle}{\|v_{i, \theta}\|_2 \cdot \|v_{i, \xi}'\|_2}  - \cdot \frac{2 \langle v_{i, \theta}, v_{i, \xi}'' \rangle}{\|v_{i, \theta}\|_2 \cdot \|v_{i, \xi}''\|_2} \Bigg)
\end{split}
\end{equation}
A symmetrical loss, that alternatively passes $V_1$ through the momentum encoder and $\{V_2, V_3\}$ through the online encoder is also produced, denoted as $\bar{\mathcal{L}}_{\text{sim}}(v_{\xi},  v_{\theta}',  v_{\theta}'')$. The total loss to train the detector head is then 
\begin{equation}
    \mathcal{L}_{\text{det}}(b, V_1, V_2, V_3) = \mathcal{L}_{\text{sim}}(v_{\theta},  v_{\xi}',  v_{\xi}'') + \bar{\mathcal{L}}_{\text{sim}}(v_{\xi},  v_{\theta}',  v_{\theta}'').
    \label{eq:det_loss}
\end{equation}

\subsection{Detector Head Pre-Training}
\label{app:soco:ssl_detector}
We train the detector head with a scale-aware assignment strategy for assigning proposals to FPN levels, analogous to SoCo~\cite{wei2021aligning}. 
Typical supervised object detection training uses small batch sizes due to the large number of region proposals for each image, where $K$ can be as high as 1000~\cite{ren2015faster}. In our setting we require a larger batch size to train the detector with the BYOL-style similarity loss~\cite{grill2020bootstrap}. 
We therefore require to adjust the number of proposals that power the detector head loss. In our experiments we set $K=4$, following the ablation in SoCo. The low proposal count is mitigated in self-supervised setups via the use of multiple views and long training times.

\subsection{Full Algorithm}
\label{app:soco:algorithm}
Our whole detection architecture is trained in a self-supervised manner through two losses. The first is the bounding box regression loss that trains the RPN head towards high quality proposals, and the second is the similarity loss that drives the detection head towards discriminative features. Through ablations on the loss terms, we find that the backbone benefits from only receiving training signal from $\mathcal{L}_{\text{det}}$. By jointly training all architectural components, we arrive at a total loss of:
\begin{equation}
\label{eq:loss}
    \mathcal{L}(p, t, b, V_1, V_2, V_3)  =  \mathcal{L}_{\text{RPN}}(p, t)  + \mathcal{L}_{\text{det}}(b, V_1, V_2, V_3).
\end{equation}
During training, we perform stochastic gradient descent to minimize this loss.

\section{Experiments} 
\label{sec:experiments}
We aim to evaluate pre-trained models in terms of the quality of their transferable representations. Firstly, following previous work~\cite{wei2021aligning}, we consider multiple downstream object detection and instance segmentation benchmarks that include common objects and additionally consider object detection under an application specific (autonomous driving) setting. Inspired by the empirical findings of the first experiment, we consider few-shot object detection, a challenging small-scale data setting, with only limited examples. In addition, ablation studies are conducted to better understand credit assignment and constituent component attributions. Our experimental setup is thus designed to answer the explicit questions, posed previously in Sec.~\ref{sec:intro}.

\subsection{Implementation}
\subsubsection{Selective Search}
Prior to self-supervised pre-training, we generate unsupervised object proposals using the heuristic selective search algorithm~\cite{uijlings2013selective}. The algorithm relies on colour and texture similarities to segment images of natural scenes into regions, which are subsequently merged to form the basis of the output bounding boxes. Default parameters are used in this process, with minimum segment size defined by $\texttt{min\_size} = 10$, output segment size and number by $\texttt{scale} = 500$, and the Gaussian kernel size for smoothing by $\sigma = 0.9$. Each bounding box $b = \{x, y, w, h\}$ is represented by its centre coordinate $(x, y)$ and its width and height $(w, h)$. We keep only boxes that satisfy $\frac{1}{3} \leq \frac{w}{h} \leq 3$ and $0.3 \leq \frac{\sqrt{wh}}{\sqrt{WH}} \leq 0.8$, where $W$ and $H$ are the height and width of the image.

\subsubsection{View Augmentation}
\label{sec:exp:aug}
During training, an input image $x$ is augmented in order to define three different views; $V_1, V_2, V_3$. View $V_1$ is resized to $224{\times}224$, $V_2$ is a random crop of $V_1$ with a randomly selected scale in the range [0.5, 1.0]. View $V_3$ is a downsampled version of $V_2$, with resulting size $112{\times}112$. The bounding boxes associated with each view are similarly scaled and shifted according to the above transformations. For each view we also apply colour jitter and blurring as in BYOL~\cite{grill2020bootstrap}. The augmentations described enables learning of scale- and translation-invariances, in a similar fashion to SoCo~\cite{wei2021aligning}.

\subsubsection{Region Proposal Network}
The region proposal network (RPN) is trained by regressing to the bounding boxes produced by selective search. At each iteration of training, $K$ random boxes are selected for each image. From that image, the feature pyramid network backbone generates a set of features $\{p2, p3, p4, p5\}$ that form the input to the RPN. While it is common to also include the final pyramid features $p6$, the images used in our self-supervised pre-training, of size $224{\times}224$, are deemed too small for the resolution of $p6$ to be useful. The RPN generates a set of anchors at each level of the feature pyramid, with areas of $\{24^2, 48^2, 96^2, 192^2\}$, respectively, and at three aspect ratios; $\{0.5, 1.0, 2.0\}$. Anchors are matched with target bounding boxes and two losses are used to update the network parameters, a classification log loss and a smooth $\mathcal{L}_1$ loss. When using RPN proposals in the detector head loss, we select the top $64$ proposals before non-maximum suppression is applied, and finally the top $K$ proposals are output.

\subsection{Training}
\label{sec:experiments:training}

We train ADePT under both the aforementioned ``separate'' and ``joint'' training strategies. We train each model for $100$ epochs using 8${\times}$ NVIDIA V100 GPUs. Training takes ${\sim}3$ days and ${\sim}4.5$ days on ImageNet~\cite{deng2009imagenet} for each strategy, respectively.
\\
\noindent \textbf{Separate Training}
We pre-train only the RPN component of the network. The feature extractor (backbone + FPN) and detector head are initialized from a SoCo baseline, pre-trained for 100 epochs. We use the LARS optimizer with a learning rate of $0.1$ annealed to $0$ with a cosine scheduler for $100$ epochs.

\noindent \textbf{Joint Training}
The joint training setup uses the RPN loss described in Eq.~\ref{eq:rpn_loss} and the detector loss described in Eq.~\ref{eq:det_loss}, that pre-trains the feature extractor and detector head. This strategy enables self-supervised training of the entire multi-stage architecture including: feature extractor, detector head and RPN from random initialization. We use the LARS optimizer with a learning rate of $0.1$ annealed to $0$ with a cosine scheduler for $100$ epochs. The RPN model layers are updated using the $\mathcal{L}_{\text{RPN}}$ loss term (Eq.~\ref{eq:rpn_loss}), exclusively.

\subsection{Transfer Learning Evaluation}
\label{sec:experiments:transfer}
\subsubsection{Experimental Setup}
Our first experimental setting follows~\cite{wei2021aligning}. For fair comparison, all SSL methods are pre-trained using ImageNet~\cite{deng2009imagenet} under identical settings. We examine the transferablity of the learned representations by fine-tuning pre-trained weights on the relevant datasets for corresponding downstream tasks. We fine-tune with stochastic gradient descent and a batch size of 16 split across 8 GPUs. 

We compare our proposed approach with recent alternative SSL methods. Among the considered baselines, three representative baselines\footnote{For these three baseline SSL methods, the results are achieved by directly running the given source code under the corresponding experimental setups for fair comparison.} are BYOL~\cite{grill2020bootstrap}, a seminal SSL baseline based on an online-target encoder architecture and image-level contrastive loss; SwAV~\cite{caron2020unsupervised} based on contrastive learning and clustering; SoCo~\cite{wei2021aligning}, the state-of-the-art. Both SoCo and ADePT leverage a BYOL-style training strategy in terms of the online-target encoders setup and instance-wise discrimination pretext task (contrastive learning). SwAV serves as an alternative contrastive learning baseline in contrast to BYOL. In addition, we quote further common baselines reported in previous work~\cite{wei2021aligning, han2021soda10m}, such as MoCo~\cite{he2020momentum}, SimCLR~\cite{chen2020simple}, DenseCL~\cite{wang2021dense}, and DetCo~\cite{xie2021detco}.

We firstly consider the \textbf{MS COCO} dataset~\cite{lin2014microsoft} and use the~\texttt{train2017} subset, containing ${\sim}118$k images for training. We train a Mask R-CNN~\cite{he2017mask} detector with a ResNet50-FPN backbone for $90000$ iterations. Eighty object categories are annotated with both bounding box and instance segmentation labels, allowing evaluation of the object detection and instance segmentation tasks. Further, we simulate two \emph{label-scarce} settings under this dataset and additionally train using $10\%$ and $1\%$ of the training set respectively, for $9000$ iterations in each case. The transfer learning performance of the learned representations is then evaluated using the MS COCO \texttt{val2017} image subset. We report the metrics $\text{AP}_{bb}$ for object detection and $\text{AP}_{mk}$ for instance segmentation, under the MS COCO 1${\times}$ schedule.

We next consider the \textbf{SODA10M} dataset~\cite{han2021soda10m} which consists of driving scenes, captured with vehicle dash cameras. The dataset contains six object classes with bounding box annotations, namely: \{car, truck, pedestrian, tram, cyclist, tricycle\} where the training set consists of only $5000$ images captured from a single city in clear weather, during the day. We again train a Cascade R-CNN object detector~\cite{cai2018cascade} with ResNet50-FPN backbone for $3726$ iterations on the full (labeled) training set of $5000$ images. We then evaluate using the 5000 image validation set where varying weather conditions, cities and periods of the day are present, making for a challenging object detection test with potential domain shifts. 

\begin{table}[t]
\centering
\resizebox{1.0\textwidth}{!}{
\begin{tabular}{l|l|r|c|c|c|c|c|c|c}
\hline
\multicolumn{3}{c|}{Pre-training} &  \multicolumn{2}{c|}{COCO} & \multicolumn{2}{c|}{COCO 10\%} & \multicolumn{2}{c|}{COCO 1\%} & SODA10M \\\hline
Method & Architecture & Epochs & AP$_{bb}$ & AP$_{mk}$ & AP$_{bb}$ & AP$_{mk}$ & AP$_{bb}$ & AP$_{mk}$ & AP$_{bb}$ \\
\hline
Scratch & R50-FPN & - & 26.4  & 34.0 & - & - & - & - & 25.4\\
Supervised & R50-FPN & 90 & 38.2  & 35.4 & - & - & - & - & 32.9\\
MoCo~\cite{he2020momentum} & R50-FPN & 200 & 38.5 & 35.1 & - & - & - & - & 32.3 \\
SimCLR~\cite{chen2020simple} & R50-FPN & 200 & - & - & - & - & - & - &32.8 \\
SwAV~\cite{caron2020unsupervised} & R50-FPN & 200 & 38.5 & 35.4 & - & - & - & - & 33.9 \\
BYOL~\cite{grill2020bootstrap} & R50-FPN & 300 & 40.4 & 37.2 & - & - & - & - & - \\
DenseCL~\cite{wang2021dense} & R50-FPN & 200 & 40.3 & 36.4 & - & - & - & -&34.3 \\
DetCo~\cite{xie2021detco} & R50-FPN & 200 & 40.1 & 36.4 & - & - & - & - & 34.7 \\\hline\hline
BYOL~\cite{grill2020bootstrap} & R50-FPN & 1000 & 40.6 & 36.7 & 24.5 & 23.2 & 13.0 & 12.5 & 41.4 \\
SwAV~\cite{caron2020unsupervised} & R50-FPN & 800 & 40.1 & 36.4 & 24.3 & 23.1 & 12.9 & 12.7 & 42.4\\
SoCo~\cite{wei2021aligning} & R50-FPN-Head & 100 & 41.8 & 37.3 & 28.2 & 25.5 & 15.0 & 13.9 & 43.3 \\
ADePT (separate) & R50-FPN-RPN-Head & 100 & \textbf{41.9} & \textbf{37.6} & \textbf{28.6} & \textbf{25.9} & \textbf{15.3} & 14.1 & 43.2\\
ADePT (joint) & R50-FPN-RPN-Head & 100 & 41.8 & 37.5 & \textbf{28.6} & 25.8 & \textbf{15.3} & \textbf{14.2} & \textbf{43.6} \\
\hline
\end{tabular}
}
\caption{Pre-trained ResNet50 models evaluated on object detection (AP$_{bb}$ / AP50$_{bb}$) and instance segmentation (AP$_{mk}$) tasks using MS COCO and SODA10M. All methods are pre-trained on ImageNet. For the upper table, we directly quote performance reported in the cited articles. A Mask R-CNN detector is used in each case. The lower table reports our experimental results, where the detector is Mask R-CNN for MS COCO and Cascade R-CNN for SODA10M, respectively.}
\label{tab:detection_results}
\end{table}

\subsubsection{Results} 
The performance is reported in \mbox{Table}~\ref{tab:detection_results}. We observe that our separately trained model consistently offers moderate improvements upon SoCo and other baselines in both object detection and instance segmentation metrics, when evaluated using MS COCO. We attribute improvements in both tasks to predominantly be due to our reduced localization error and model sensitivities to related concepts. The joint training strategy provides overall competitive performance with SoCo on MS COCO, but achieves state-of-the-art performance on the challenging SODA10M benchmark. Generally, with smaller fine-tuning datasets (MS COCO), the performance gap between ADePT and SoCo becomes larger. In comparison with the performance of SoCo, this partially answers our \textbf{Q1} and we conclude that RPN pre-training provides a practical solution towards improving SSL performance. To provide further insight towards \textbf{Q1}, we provide an ablation study on the error analysis in Sec.~\ref{sec:experiments:ablation}.

We note that SwAV and BYOL can only achieve comparable performance under much increased epoch count training regimes. When equal pre-training epoch counts are considered, SoCo and ADePT outperform these image-level pre-training baselines by learning object-level features for the relevant downstream tasks. Additionally it is well understood that MS COCO is characterized as containing a high percentage of small annotated objects~\cite{pont2015boosting}. As shown in Fig.~\ref{fig:qualitative_coco}, smaller objects may be regarded as harder to detect on average and the additional benefits, brought about by pretext task alignment (\ie~RPN pre-training), can bring detection performance improvements in such situations. This observation helps to resolve \textbf{Q2}: the alignment between the pretext task and downstream task can bring improvement to overall performance.

\mbox{Table}~\ref{tab:detection_results} results might allow one to argue that absolute performance gains, observed when pre-training the RPN, are small. However MS COCO and SODA10M are large datasets (\eg~$1\%$ of MS COCO contains \mbox{${\sim}1.18K$} images). We thus conjecture that fine-tuning all model weights, in the investigated settings, may benefit SoCo and other SSL baselines disproportionately. With large-scale labeled data available, the RPN component (randomly initialized for SoCo and other baselines) can be deemed easier to fine-tune in comparison with both feature extractor and detector heads, due to the relatively small number of learnable parameters. To explore this point further, we design a following experiment (Sec.~\ref{sec:experiments:few_shot}), to better understand the contribution of RPN pre-training.

\subsection{Few-shot Evaluation}
\label{sec:experiments:few_shot}
\subsubsection{Experimental Setup} In addition to the transfer evaluation protocol described previously in Sec.~\ref{sec:experiments:transfer}, we further realize a new protocol to evaluate the performance of self-supervised learning under scenarios that exhibit extreme label-scarcity. The proposed protocol is inspired by the benchmark few-shot object detection evaluation convention adopted in~\cite{wang2020frustratingly,fan2021generalized}. Few-shot object detection tasks consist of a set of base classes and a set of novel classes, where base classes have abundant labels and yet novel classes are provided with only few labels. 
Without loss of generality, we take~\cite{wang2020frustratingly} as an example. There are two training stages in~\cite{wang2020frustratingly}, namely, a base training stage and a few-shot fine-tuning stage. During base training, standard supervised training is performed, and in the few-shot fine-tuning stage, the model is expected to learn to detect the novel classes given only $K$ labelled instances for each class of interest, \ie~a $K$-shot learning task. In this work, we alternatively abandon the base training stage and pre-train the model (Faster R-CNN with a ResNet50 backbone + FPN) on ImageNet~\cite{deng2009imagenet}, \ie~no labelled data are involved in the first stage. Then, following~\cite{wang2020frustratingly,fan2021generalized}, we fine-tune the model via $K$-shot training~\cite{vinyals2016matching,snell2017prototypical}. Finally we use a proxy evaluation task to assess both representation and fast adaptation capabilities of the pre-trained model weights. We note that such proxy evaluation protocols are commonly adopted in representation learning, \eg~\cite{he2020momentum,chen2020simple,dong2022residual}, where only the last layer is fine-tuned with other layers fixed. As SoCo remains \emph{unable} to pre-train an RPN component, towards fair comparison, we instead opt to fine-tune their entire model. 

We perform the proposed few-shot evaluation on the \textbf{\mbox{PASCAL} VOC} dataset~\cite{everingham2010pascal}. We follow the fine-tuning and evaluation setup in \cite{wang2020frustratingly}. We use the \texttt{trainval07+12} set with the same data splits as \cite{wang2020frustratingly} and report object detection performance on the \texttt{test07} set, using the standard VOC metric; \ie~mAP50.

\begin{table*}[t]
\centering
\resizebox{1.0\textwidth}{!}{
\begin{tabular}{c|l|l|c|c|c|c|c|c|c|c|c|c|c|c|c|c|c}
\hline
{} & \multicolumn{2}{c|}{Pre-training} &  \multicolumn{5}{c|}{Split 1} & \multicolumn{5}{c|}{Split 2} & \multicolumn{5}{c}{Split 3} \\
\hline
{} & Method & Architecture & $K{=}1$ & $K{=}2$ & $K{=}3$ & $K{=}5$ & $K{=}10$ & $K{=}1$ & $K{=}2$ & $K{=}3$ & $K{=}5$ & $K{=}10$ & $K{=}1$ & $K{=}2$ & $K{=}3$ & $K{=}5$ & $K{=}10$\\
\hline
\multirow{3}{*}{\rotatebox{90}{Novel}} & SoCo~\cite{wei2021aligning} & R50-FPN-Head & 0.56 & 1.00 & 0.94 & 1.91 & 2.96 &  0.19 & 0.73 & 1.26 & 2.29 & 1.68 & 0.12 & 1.52 & 1.22 & 1.48 & 3.42 \\
{} & ADePT (sep) & R50-FPN-RPN-Head & 1.60 & \textbf{2.32} &    \textbf{1.64} & 1.17 &  \textbf{3.52} & 1.69 &  \textbf{1.94} & 2.07 &  1.85 &  3.74 & \textbf{3.03} & \textbf{2.13} & 1.42 & \textbf{3.56} & 3.56  \\
{} & ADePT (joint) & R50-FPN-RPN-Head  & \textbf{2.14} &
1.23  & 1.24  & \textbf{1.93} & 2.91  & \textbf{2.53} & 0.59  & \textbf{2.80}  & \textbf{5.55}  & \textbf{4.18} & 1.07  &   1.47  & \textbf{1.43}  & 2.05 & \textbf{5.72}\\
\hline
\multirow{3}{*}{\rotatebox{90}{All}} & SoCo~\cite{wei2021aligning} & R50-FPN-Head & 1.14 & 0.99 & 1.26 &    1.50 &  2.89 &  0.68 &  1.53 & 1.12 & 1.49 &    3.23 & 0.96 & 1.23 & 2.08 & 1.83 & 4.45
\\
{} & ADePT (sep) & R50-FPN-RPN-Head & \textbf{1.72} & \textbf{1.35} & \textbf{1.63} & 2.25 & \textbf{4.53} & 1.61 & \textbf{1.87} & 1.91 & 4.42 & 3.70 & \textbf{1.60} & \textbf{1.91} & \textbf{2.51} &    \textbf{2.42} & 4.45 \\
{} & ADePT (joint) & R50-FPN-RPN-Head & 1.62 & 1.20 & 1.41 & \textbf{2.59} &    4.20 & 2.04 & 1.60 & \textbf{2.35} &    \textbf{4.52} & \textbf{4.31} & 0.68 &  1.07 & 1.39 &   1.68 & \textbf{4.60}
\\
\hline
\end{tabular}
}
\caption{Few-shot object detection performance (mAP50) on the PASCAL VOC dataset. Our approach consistently outperforms the SoCo baseline by a large margin. $K$ denotes $K$-shot detection task.
}
\label{tab:lowdata_results}
\end{table*}

\subsubsection{Results} In Table~\ref{tab:lowdata_results} we report our few-shot learning results. We follow the same protocol as~\cite{fan2021generalized}, where we evaluate the few-shot fine-tuned model weights on three different splits of novel classes and all classes (novel classes and base classes). The detector is a Faster R-CNN and the model weights are pre-trained using ImageNet in a self-supervised fashion. We observe that ADePT can consistently outperform SoCo in the considered $K = \{1, 2, 3, 5, 10\}$ settings. Note, the unsupervised pre-training is performed on ImageNet, \ie~VOC data are only seen in the few-shot fine-tuning stage. This proxy evaluation evidences that ADePT has the ability to learn more useful representations than SoCo and shows better potential for fast-adaptation tasks, with few labels. With small $K$, the self-supervised pre-training in effect plays a critical role in the few-shot learning through the provision of strong initialization. 
In comparison with Table~\ref{tab:detection_results}, we highlight that the performance gain with RPN pre-training becomes yet larger. Furthermore, in 
Table~\ref{tab:lowdata_results}, the performance gain of RPN pre-training is larger, specifically, for the novel classes, \emph{c.f.}~the base classes, where only base classes are seen during the fine-tuning stage.
This suggests that RPN pre-training aids the learning of representations that can quickly adapt to new tasks.
Finally, we can provide some initial evidence towards answering \textbf{Q3}: RPN pre-training, with architectural alignment, is capable of bringing benefit to label-efficient regimes. 

Based on the results in Table~\ref{tab:detection_results} and Table~\ref{tab:lowdata_results}, we note that, although both of our training strategies can show improvement over SoCo, performance between our \emph{separate} and \emph{joint} training models is largely comparable in the majority of the investigated scenarios. The former strategy requires lower computation-cost, which may lead to it being favoured in practical situations. 

\subsection{Ablation Studies}
\label{sec:experiments:ablation}

\begin{table}[th]
\centering
\begin{tabular}{l| c| c| c| c| c| c| c}
\hline
Method & Cls & Loc & Dupe & Bkg & Miss & FalsePos & FalseNeg \\
\hline
SoCo & 3.39 & 6.45 &  \textbf{0.21} & \textbf{4.15} & 6.89 & \textbf{16.81} & 14.57 \\
ADePT (sep) & \textbf{3.18} & \textbf{6.28} & 0.22 & 4.35 & 6.61 & 17.40 & \textbf{13.59} \\
ADePT (joint) & 3.31 & 6.44 & 0.22 & 4.40 & \textbf{6.54} & 17.45 & 13.75 \\
\hline
\end{tabular}
\caption{Analysis of stratified errors. ADePT (separate training) exhibits the lowest localization error.}
\label{tab:errors}
\end{table}

\subsubsection{Stratification of Error Types}
Object detection is a complex task with a multitude of potential error types. Such errors can be categorised and quantified by tools such as TIDE~\cite{bolya2020tide}. In Table~\ref{tab:errors}, we stratify these errors for the detection models fine-tuned on the full MS COCO dataset by TIDE. The error types, from left to right are; classification (\emph{Cls}), localization (\emph{Loc}), duplicates (\emph{Dupe}), background objects (\emph{Bkg}), missed objects (\emph{Miss}), false positives (\emph{FalsePos}), and false negatives (\emph{FalseNeg}).

We observe that ADePT achieves the lowest localization error, with ADePT (sep) achieving a significant reduction with respect to the baselines. This is also visualized in Fig.~\ref{fig:cls_and_loc_errors}. The classification errors of these models are additionally reduced compared to SoCo. Notably, ADePT has low error in the ``missed objects'' category along with low ``false negative'' error, which may be interpreted intuitively in terms of the rate of successfully locating \emph{all} objects in images. Conversely, the ``background object'' error and ``false positive'' rates are higher, which we conjecture to illustrate that the greater reliance on selective search proposals produces a model that is prone to false positives, finding objects where they do not exist. We conclude that RPN pre-training can reduce the localization error, which corroborates our affirmative answer to \textbf{Q1}.

\begin{table}[th]
    \centering
    \resizebox{1.0\textwidth}{!}{
    \begin{tabular}{l|l|c|c|c|c|c|c|c}
        \hline
        \multicolumn{2}{c|}{ADePT Setting} & \multicolumn{2}{c|}{COCO} & \multicolumn{2}{c|}{COCO 10\%} & \multicolumn{2}{c|}{COCO 1\%} & SODA10M \\\hline
        Loss & Proposals & AP$_{bb}$ & AP$_{mk}$ & AP$_{bb}$ & AP$_{mk}$ & AP$_{bb}$ & AP$_{mk}$ & AP$_{bb}$ \\
        \hline
        $\mathcal{L}_{\text{Det}} {+} \mathcal{L}_{\text{RPN}}$ & SS {+} RPN  & \phantom{0}6.784 & \phantom{0}6.528 & \phantom{0}0.003 & \phantom{0}0.002 & \phantom{0}0.001 & \phantom{0}0.000 & 36.853\\ 
        $\mathcal{L}_{\text{Det}}$           & SS {+} RPN  & \phantom{0}0.564 & \phantom{0}0.771 & \phantom{0}0.020 & \phantom{0}0.018 & \phantom{0}0.015 & \phantom{0}0.019 & \phantom{0}0.006 \\
        $\mathcal{L}_{\text{Det}} {+} \mathcal{L}_{\text{RPN}}$ & SS        & 41.158 & 37.002 & 27.783 & 25.073 & 14.508 & 13.332 & 43.168 \\ 
        $\mathcal{L}_{\text{Det}}$           & SS        & \textbf{41.677} & \textbf{37.342} & \textbf{28.576} & \textbf{25.794} & \textbf{15.161} & \textbf{13.938} & \textbf{43.570} \\ 
        \hline
    \end{tabular}
    }
    \caption{Ablations on ADePT joint training show that the backbone benefits slightly from only receiving the detector head loss signal and that the RPN proposals tend to destabilize detector head training. SS: selective search.}
    \label{tab:ablation_results}
\end{table}

\subsubsection{Analysis of Joint Training}
\label{app:analysis:jt}

In contrast to separate training, we note that joint training can be affected by several confounding factors. Namely ADePT consists of two component losses; firstly the RPN pre-training loss, $\mathcal{L}_\text{RPN}$ (Eq.~\ref{eq:rpn_loss})  and secondly the self-supervised loss that pre-trains the feature extractor and detector head, $\mathcal{L}_\text{det}$ (Eq.~\ref{eq:det_loss}). During joint-training, the feature extractor is affected by both of these losses, while the detector head may receive proposals from either: solely selective search or selective search in conjunction with the RPN. In Table~\ref{tab:ablation_results}, we investigate how to best combine the component losses and the proposal-generation options.

There are two important findings. First, the quality of proposals plays a more important role in determining the performance. As the RPN module is learning, by inspection we find that intermediate proposals are often noisy, which introduces a destabilising effect to the detector head loss and lowers the overall learning performance. Thus, using proposals only generated by the selective search is recommended. Second, with proposals generated by the selective search, the loss signal to the feature extractor has limited effect on the performance, while a lone $\mathcal{L}_\text{det}$ offers improved performance over $\mathcal{L}_{\text{det}} {+} \mathcal{L}_{\text{RPN}}$. When using proposals generated by both the selective search and the RPN, we notice that there is a catastrophic degeneration of performance. We conjecture that in a data scarce regime, without $\mathcal{L}_{\text{RPN}}$, the phenomenon of the noisy RPN proposals mentioned above will be exacerbated.

\begin{figure}[th]
    \centering
    \begin{subfigure}[t]{0.19\textwidth}
        \centering
        \includegraphics[width=1\textwidth]{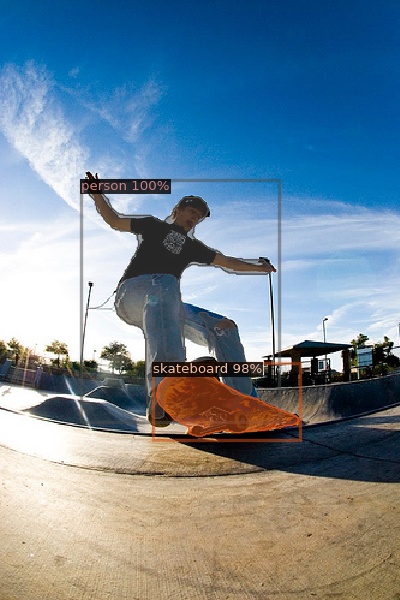}
        \includegraphics[width=1\textwidth]{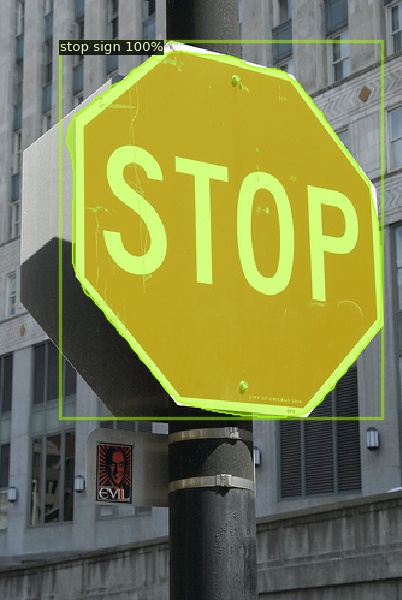}
        \caption{BYOL}
    \end{subfigure}%
    \begin{subfigure}[t]{0.19\textwidth}
        \centering
        \includegraphics[width=1\textwidth]{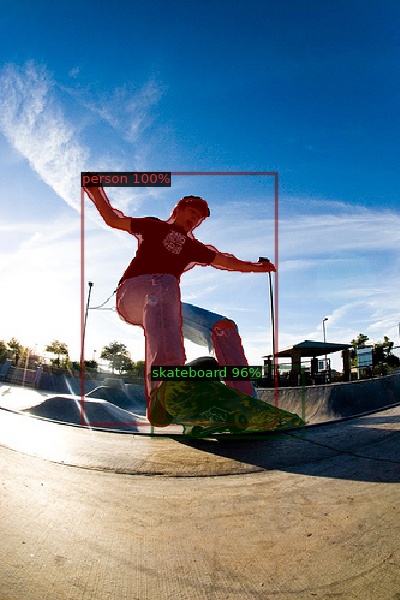}
        \includegraphics[width=1\textwidth]{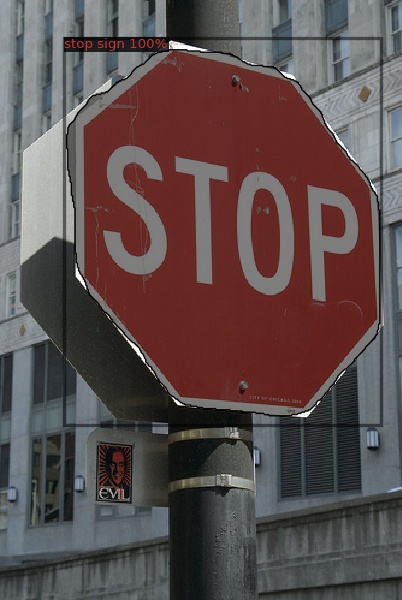}
        \caption{SwAV}
    \end{subfigure}%
    \begin{subfigure}[t]{0.19\textwidth}
        \centering
        \includegraphics[width=1\textwidth]{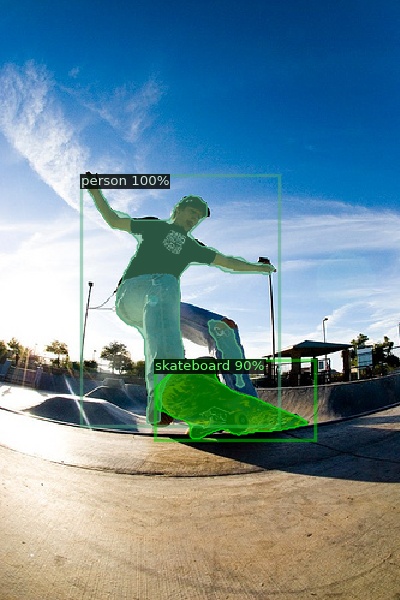}
        \includegraphics[width=1\textwidth]{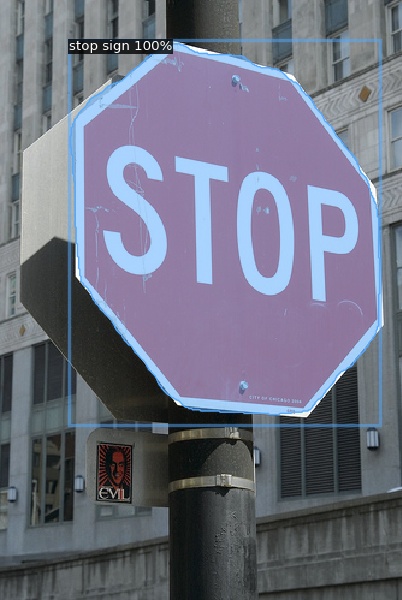}
        \caption{SoCo}
    \end{subfigure}%
    \begin{subfigure}[t]{0.19\textwidth}
        \centering
        \includegraphics[width=1\textwidth]{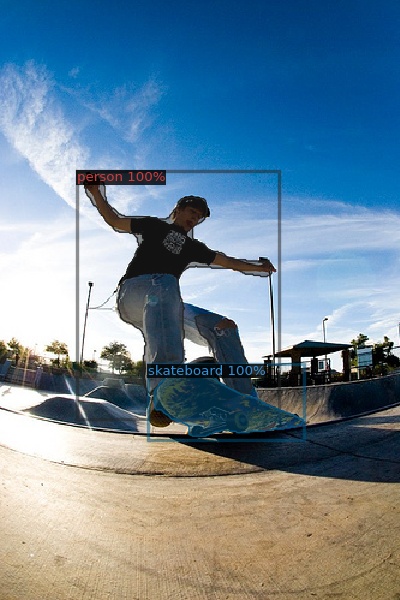}
        \includegraphics[width=1\textwidth]{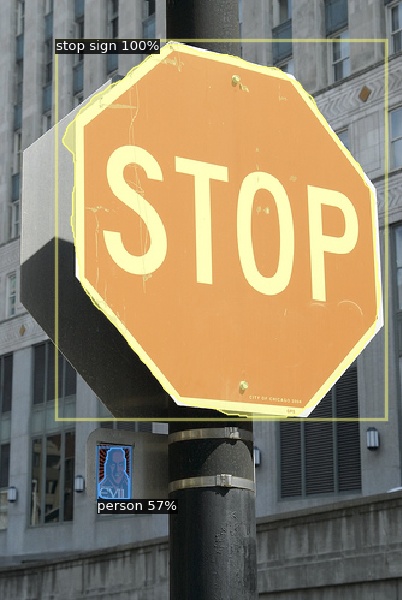}
        \caption{{\scriptsize ADePT (sep)}}
    \end{subfigure}%
    \begin{subfigure}[t]{0.19\textwidth}
        \centering
        \includegraphics[width=1\textwidth]{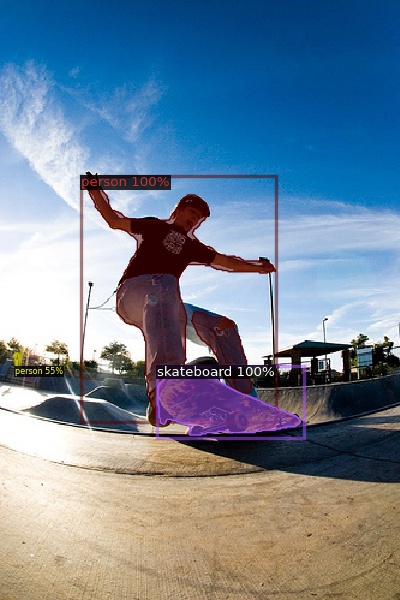}
        \includegraphics[width=1\textwidth]{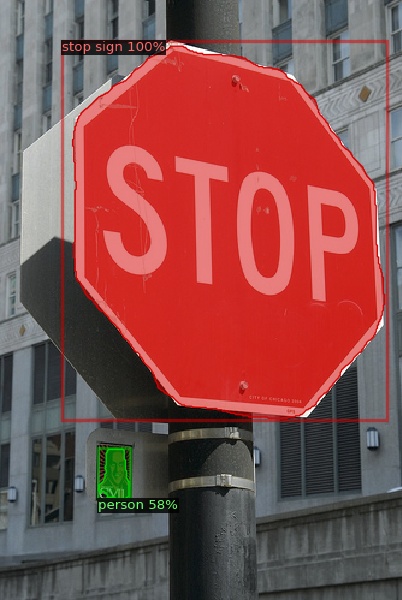}
        \caption{{\scriptsize ADePT (joint)}}
    \end{subfigure}
    \caption{Qualitative comparison on MS COCO~\cite{lin2014microsoft}. Top: our models have highest prediction confidence and ADePT (joint) additionally detects a distant person, despite difficult illumination. Bottom: all models capture the large foreground stop-sign. We further detect the ``person'' object, a small human-face image, attached to the sign. RPN pre-training can improve small object localization.}
    \label{fig:qualitative_coco}
\end{figure}

\begin{figure}[th]
    \centering
    \begin{subfigure}[t]{0.19\textwidth}
        \centering
        \includegraphics[width=1\textwidth]{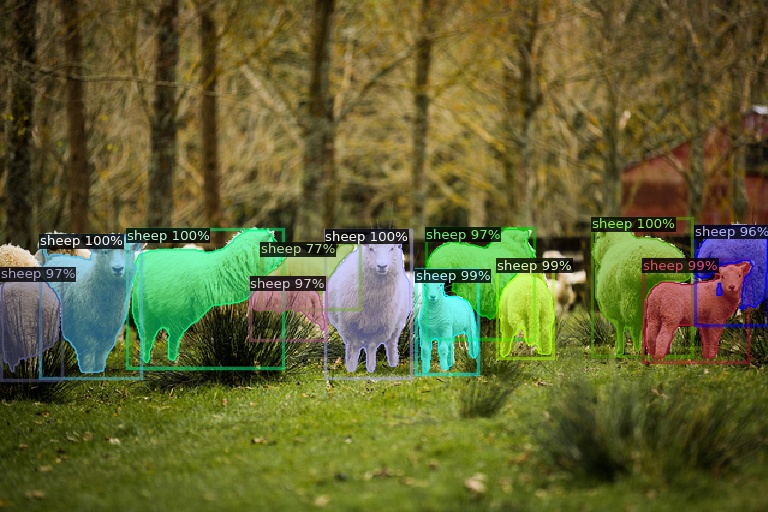}
        \caption{BYOL}
    \end{subfigure}
    \begin{subfigure}[t]{0.19\textwidth}
        \centering
        \includegraphics[width=1\textwidth]{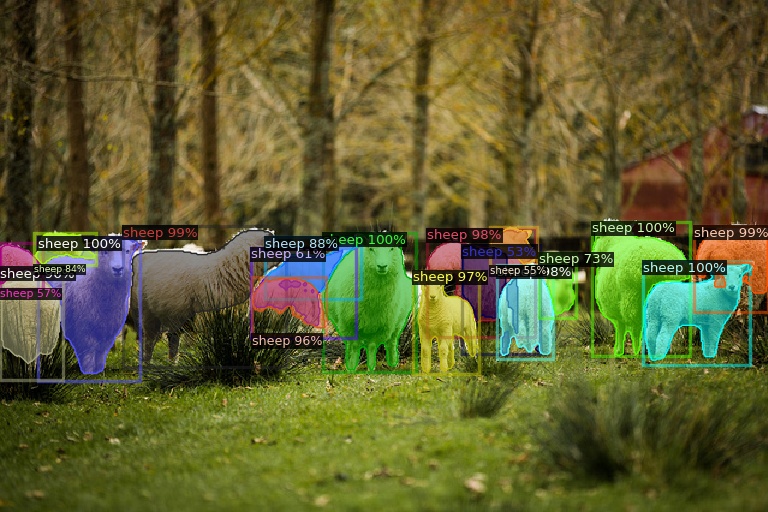}
        \caption{SwAV}
    \end{subfigure}
        \begin{subfigure}[t]{0.19\textwidth}
        \centering
        \includegraphics[width=1\textwidth]{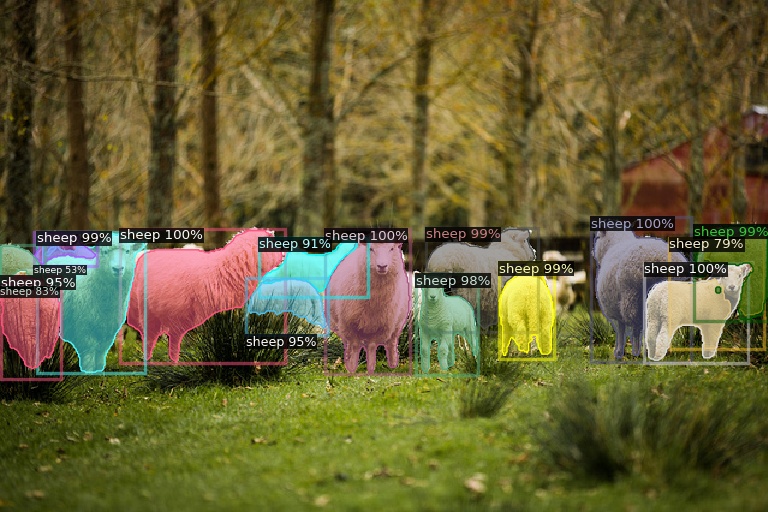}
        \caption{SoCo}
    \end{subfigure}
        \begin{subfigure}[t]{0.19\textwidth}
        \centering
         \includegraphics[width=1\textwidth]{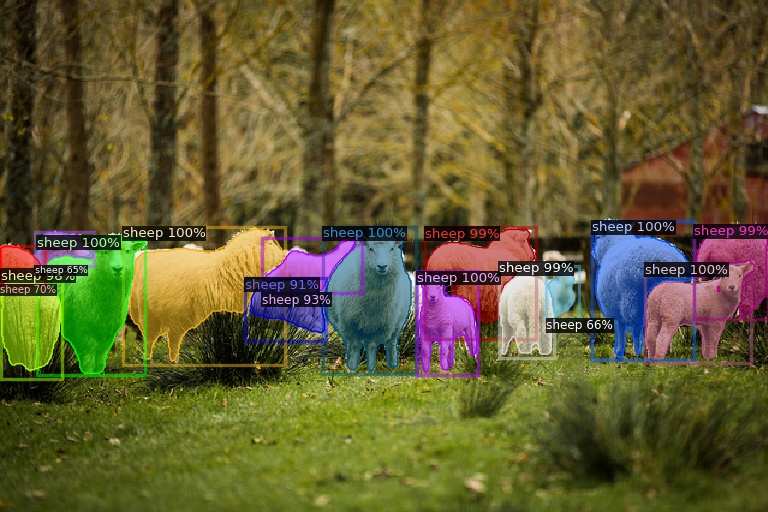}
        \caption{\scriptsize ADePT (sep)}
    \end{subfigure}
    \begin{subfigure}[t]{0.19\textwidth}
        \centering
        \includegraphics[width=1\textwidth]{figures/coco/000059/output_59_adept_separate.jpg}
        \caption{\scriptsize ADePT (joint)}
    \end{subfigure}
    \caption{Predictions using MS COCO~\cite{lin2014microsoft}. BYOL~\cite{grill2020bootstrap} fails to detect a sheep (image left).
    SwAV~\cite{caron2020unsupervised} overcounts sheep by incorrectly predicting a (single) sheep as multiple instances (image middle). SoCo~\cite{wei2021aligning} fails to detect a sheep (image middle) and incorrectly splits a single sheep in two (image right). ADePT models detect {all} instances with high confidence. \textbf{Best viewed with digital zoom}.}
    \label{fig:qualitative_coco:3}
\end{figure}

\begin{figure}[th]
    \centering
    \begin{subfigure}[t]{0.195\textwidth}
        \centering
        \includegraphics[width=1\textwidth]{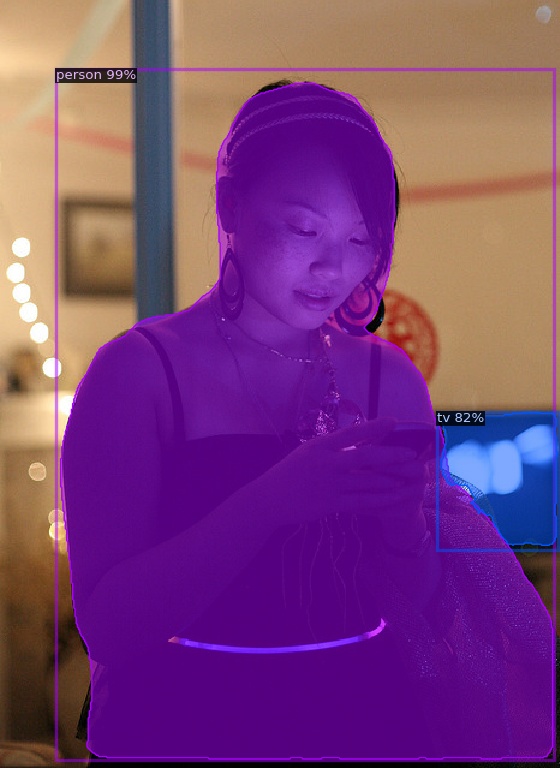}
        \caption{BYOL}
    \end{subfigure}%
    \begin{subfigure}[t]{0.195\textwidth}
        \centering
        \includegraphics[width=1\textwidth]{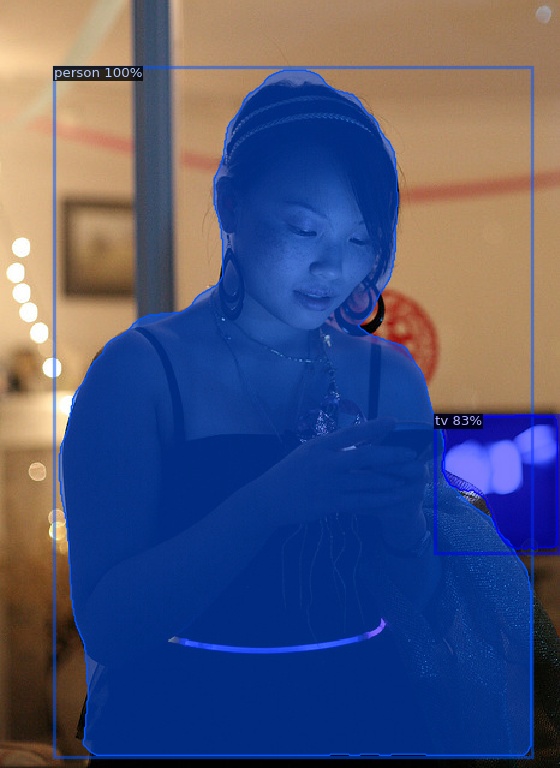}
        \caption{SwAV}
    \end{subfigure}%
    \begin{subfigure}[t]{0.195\textwidth}
        \centering
        \includegraphics[width=1\textwidth]{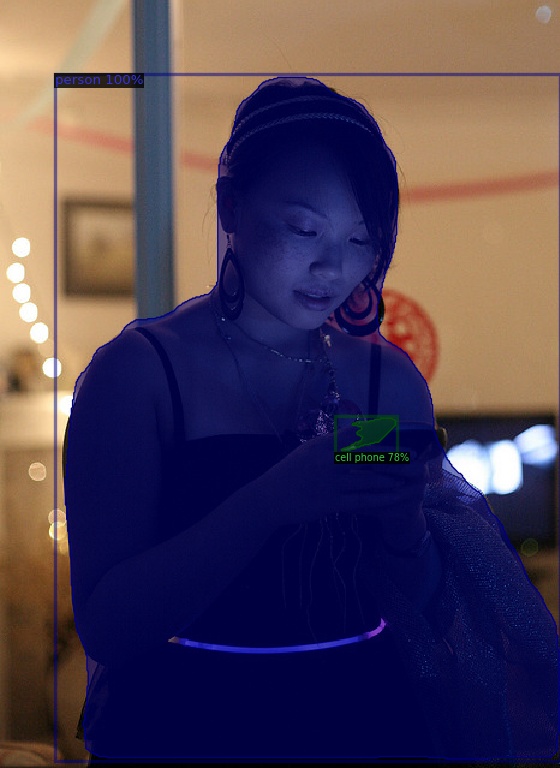}
        \caption{SoCo}
    \end{subfigure}%
    \begin{subfigure}[t]{0.195\textwidth}
        \centering
        \includegraphics[width=1\textwidth]{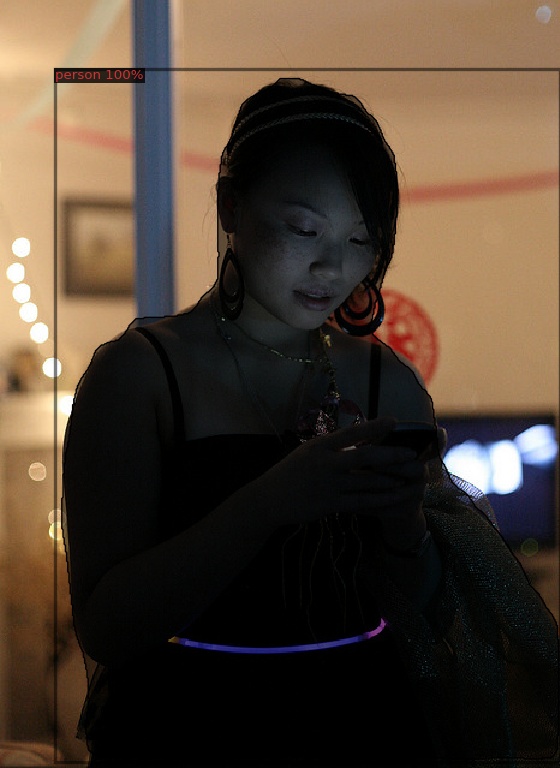}
        \caption{{\scriptsize ADePT (sep)}}
    \end{subfigure}%
    \begin{subfigure}[t]{0.195\textwidth}
        \centering
        \includegraphics[width=1\textwidth]{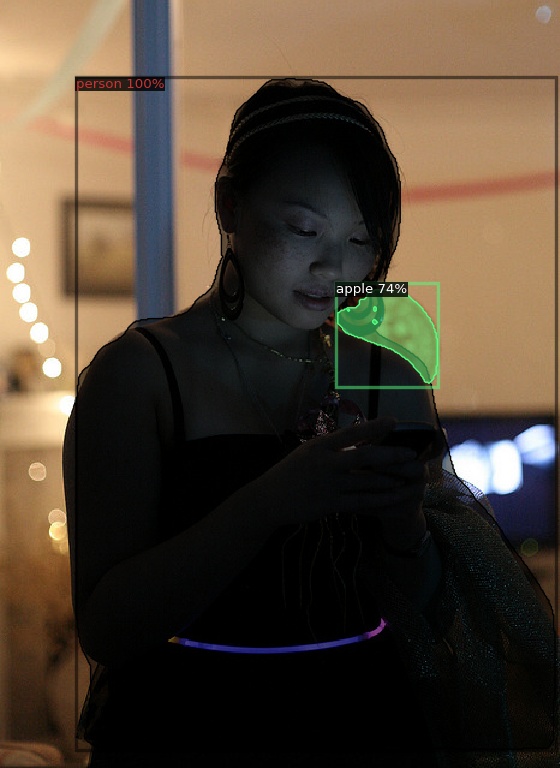}
        \caption{\scriptsize ADePT (joint)}
    \end{subfigure}
    \caption{Failure case under MS COCO~\cite{lin2014microsoft}. All methods successfully detect the person in the middle of the image, while missing additional smaller objects. BYOL and SwAV manage to detect the TV in the background under blurry imaging conditions. SoCo can detect the partially occluded phone in the hand. Our models fail to detect the TV and the phone objects and the joint training also introduces a false positive, due to the background appearance, under this challenging situation.}
    \label{fig:qualitative_coco:4}
\end{figure}

\begin{figure}[th]
    \centering
    \begin{subfigure}[t]{0.19\textwidth}
        \centering
        \includegraphics[width=1\textwidth]{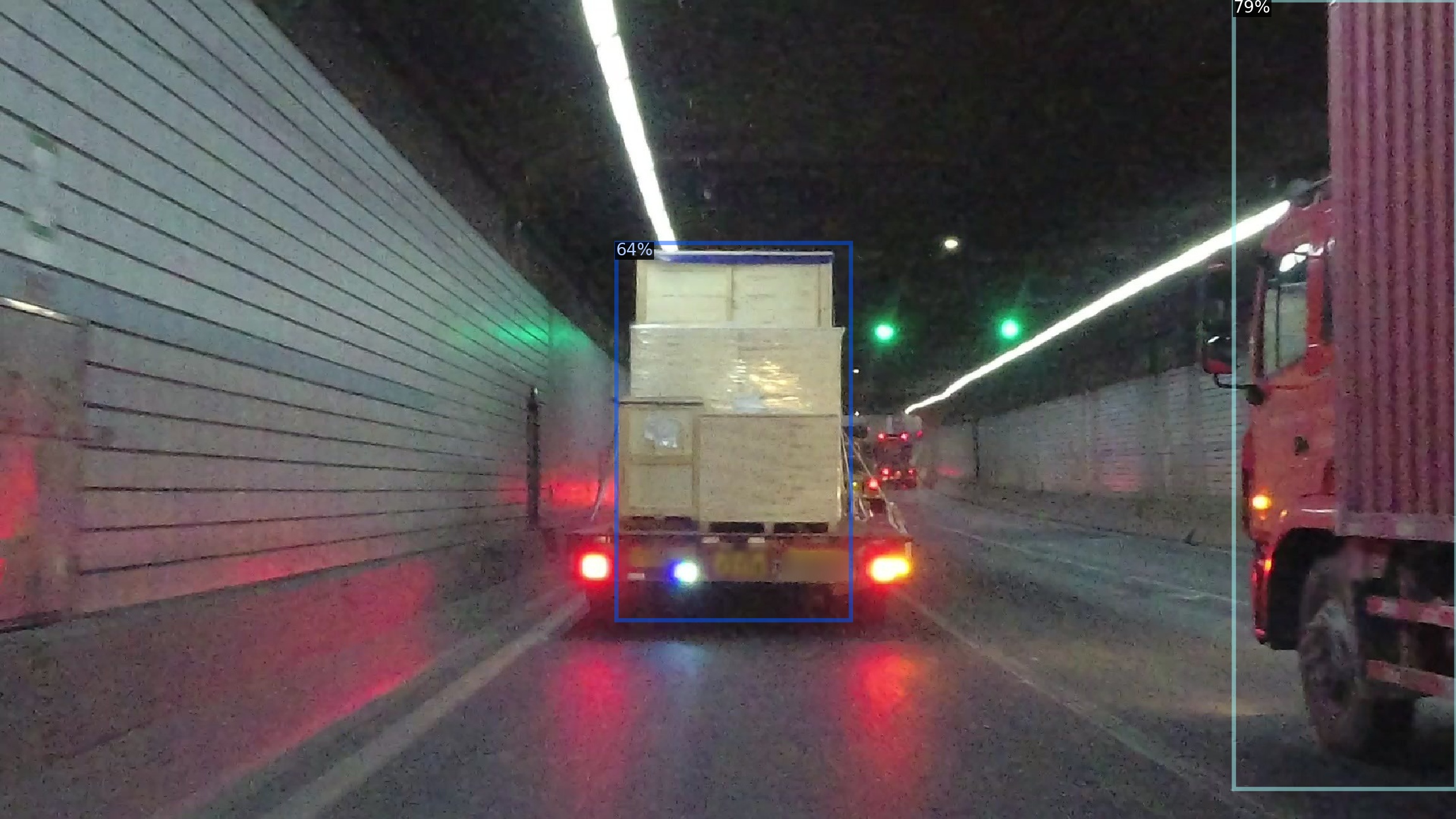}
        \includegraphics[width=1\textwidth]{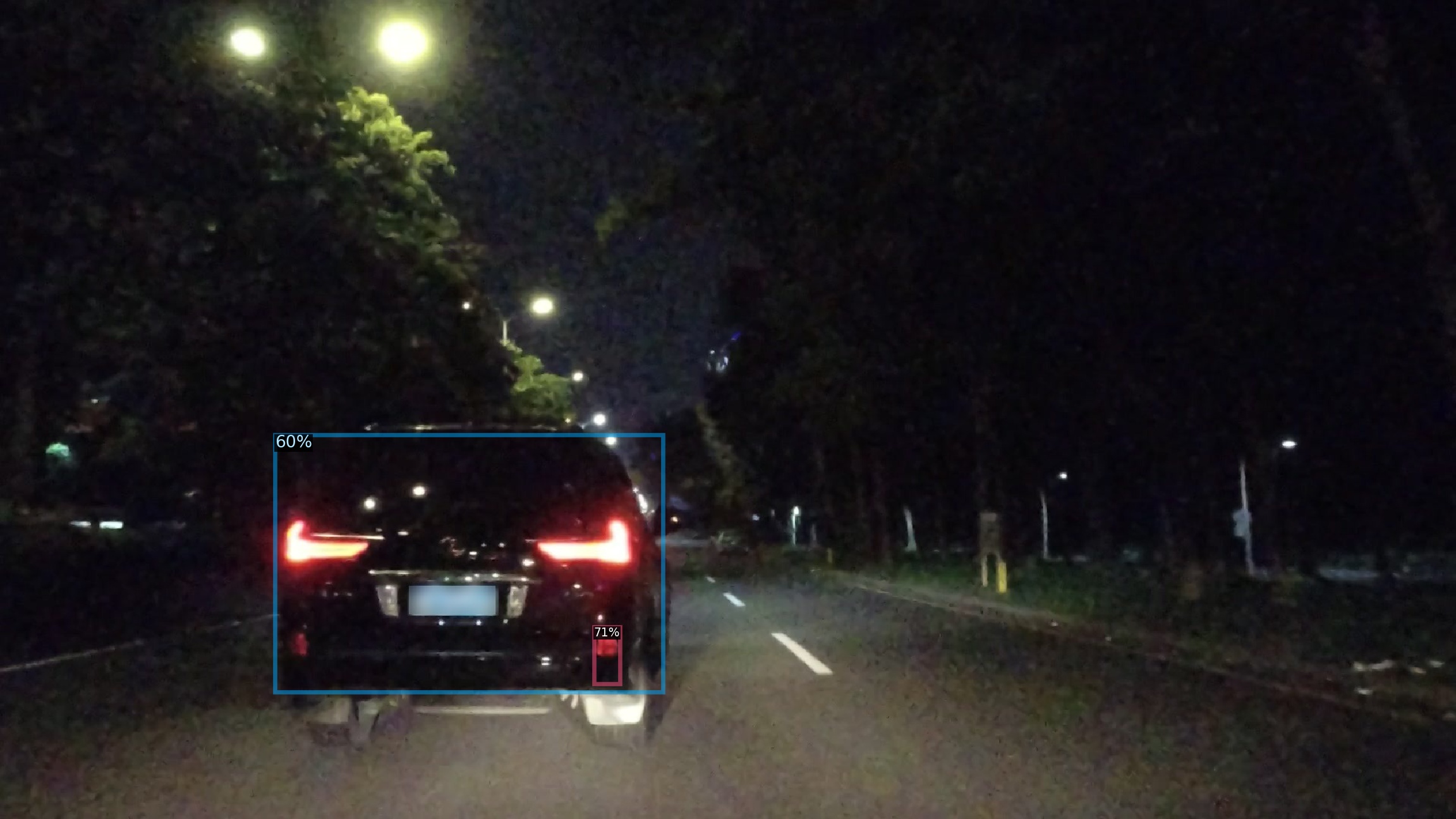}
        \caption{BYOL}
    \end{subfigure}
    \begin{subfigure}[t]{0.19\textwidth}
        \centering
        \includegraphics[width=1\textwidth]{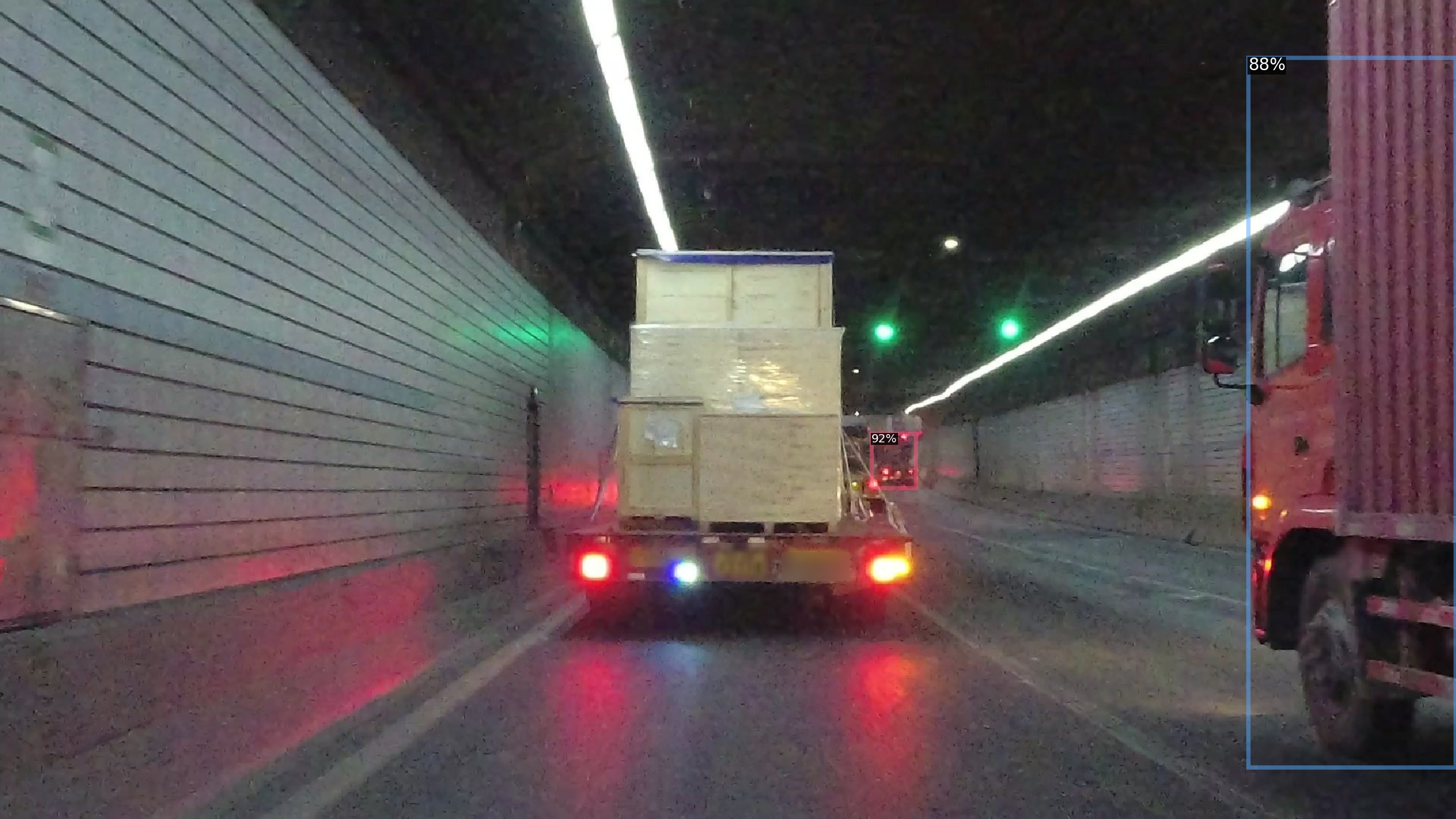}
        \includegraphics[width=1\textwidth]{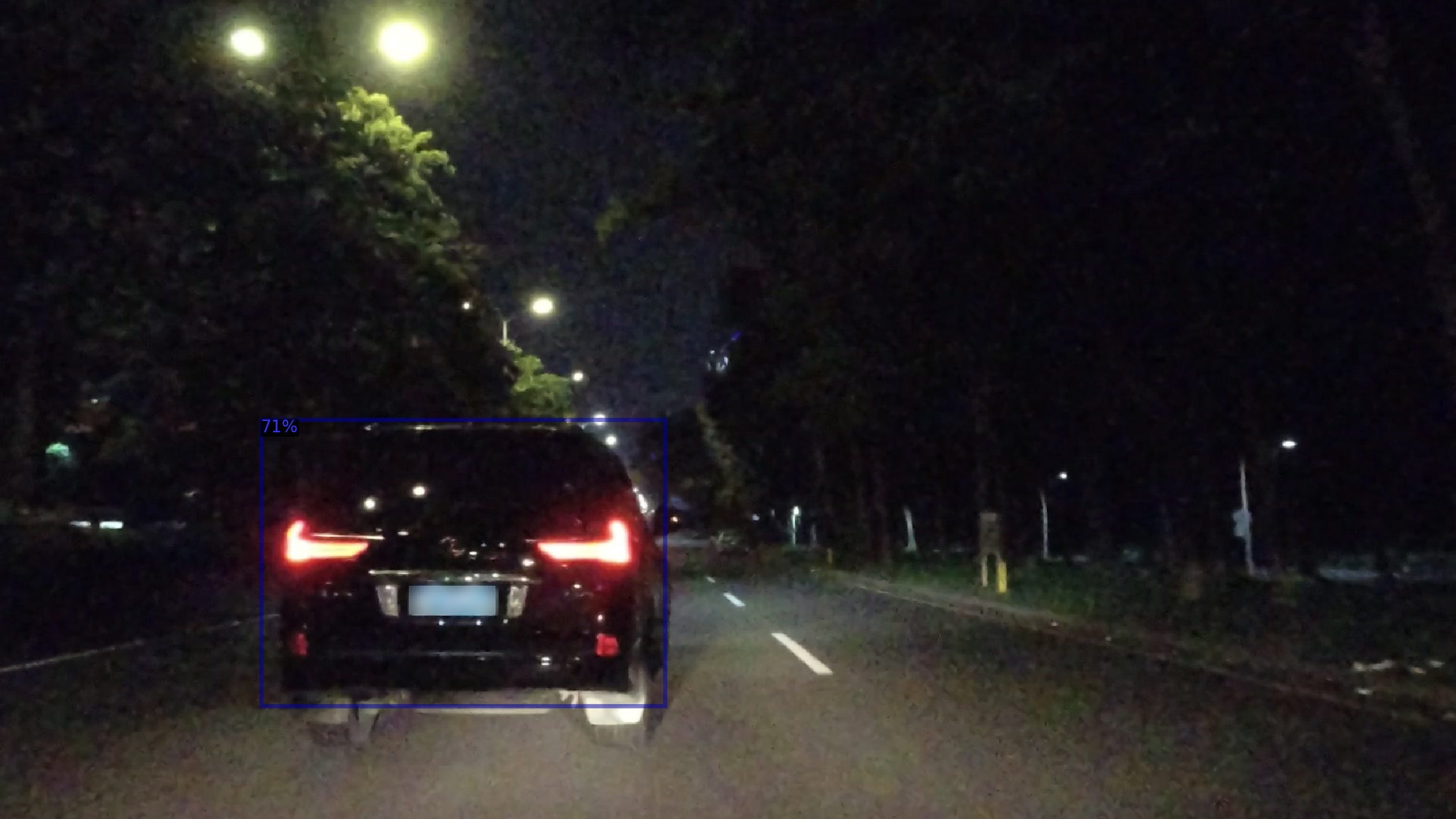}
        \caption{SwAV}
    \end{subfigure}
    \begin{subfigure}[t]{0.19\textwidth}
        \centering
        \includegraphics[width=1\textwidth]{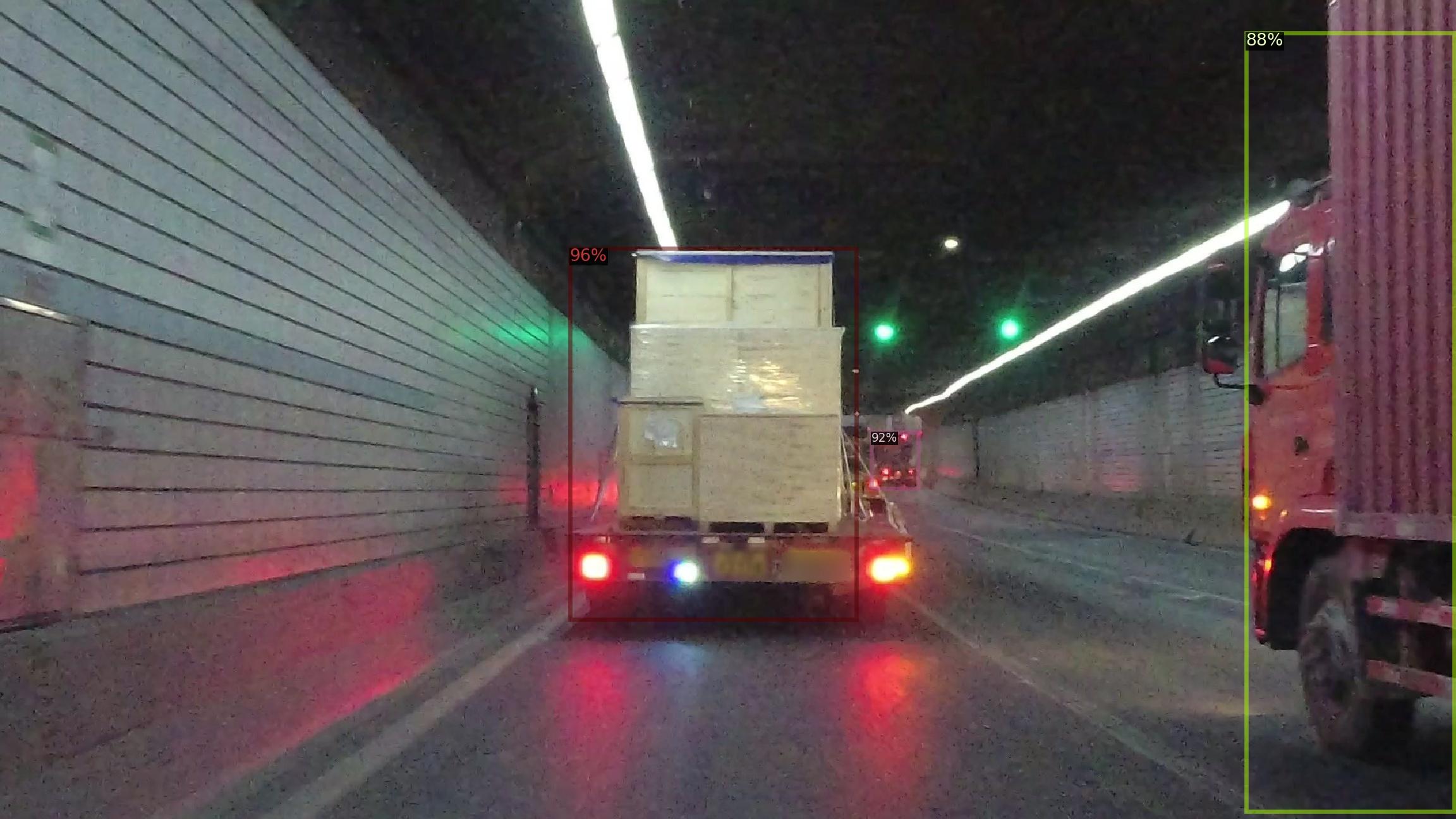}
        \includegraphics[width=1\textwidth]{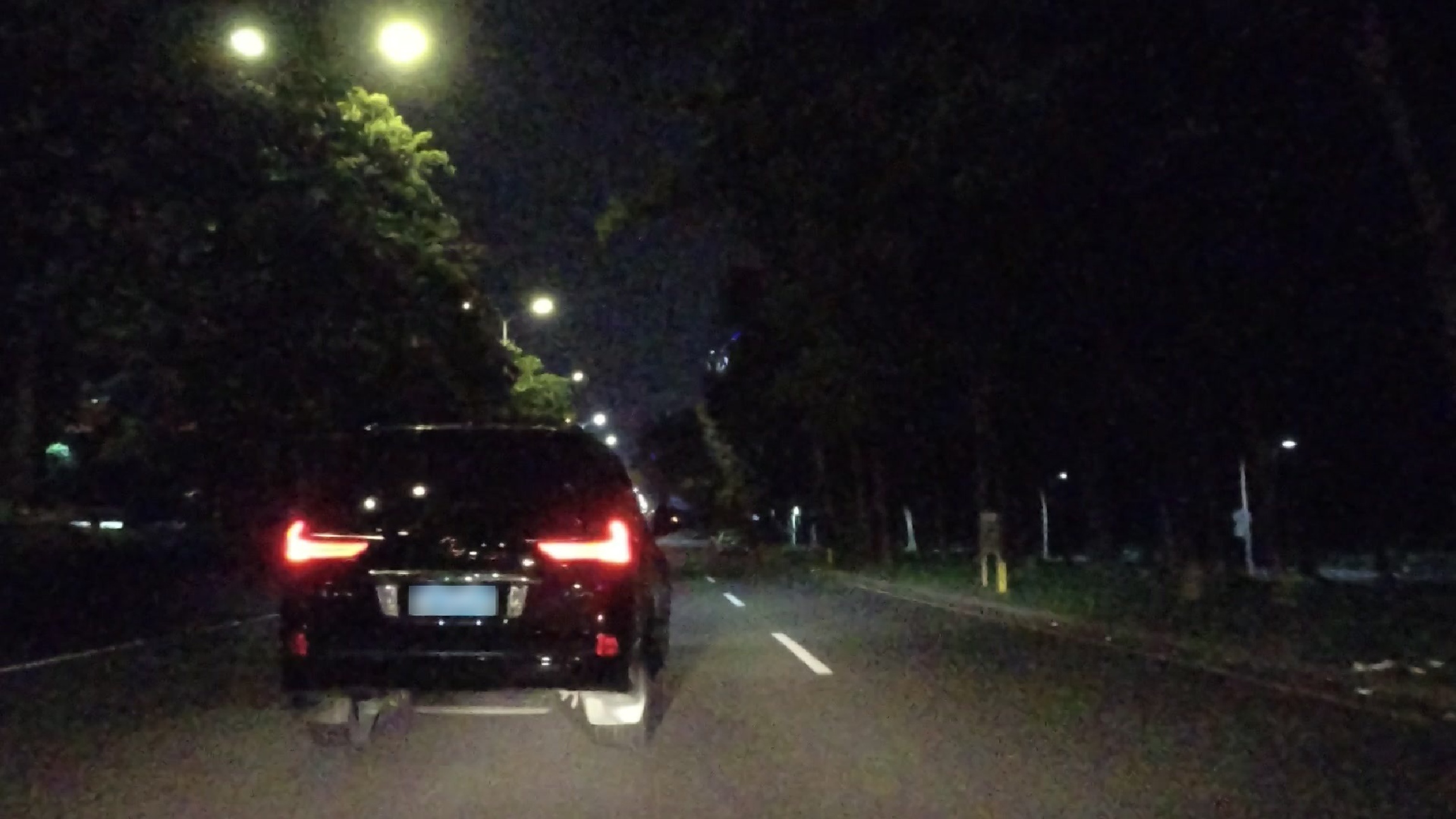}
        \caption{SoCo}
    \end{subfigure}
    \begin{subfigure}[t]{0.19\textwidth}
        \centering
        \includegraphics[width=1\textwidth]{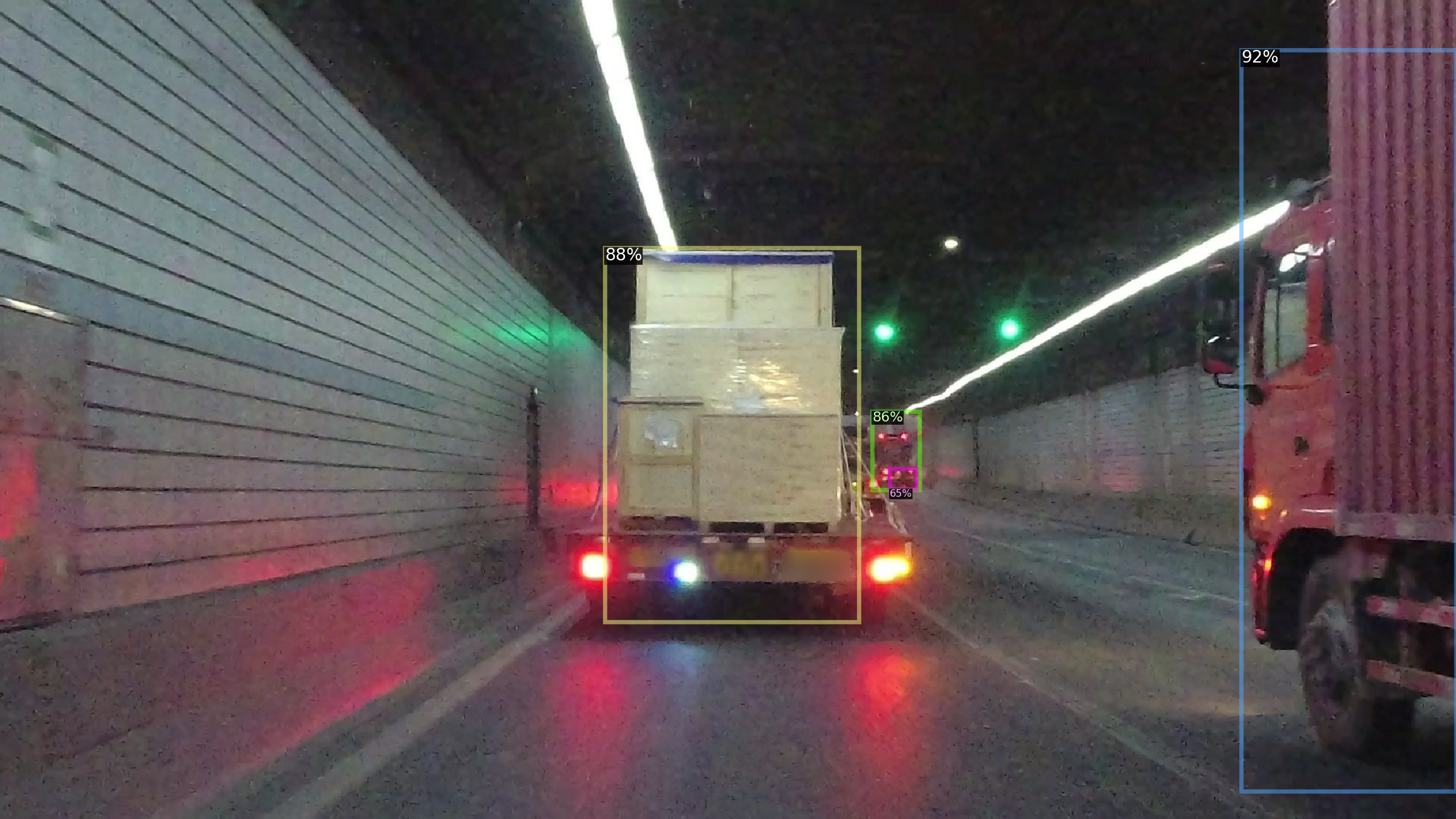}
        \includegraphics[width=1\textwidth]{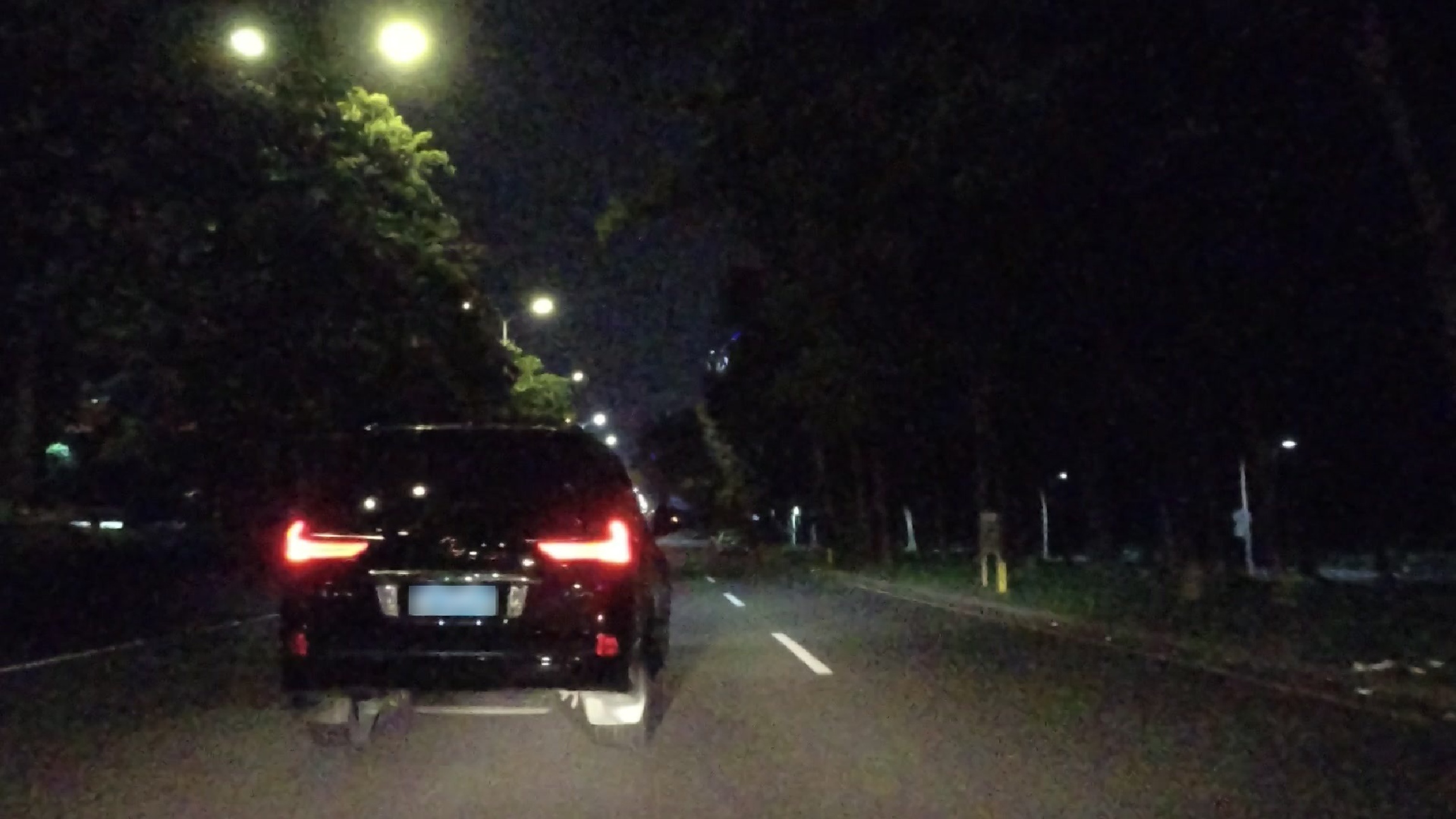}
        \caption{{\scriptsize ADePT (sep)}}
    \end{subfigure}
    \begin{subfigure}[t]{0.19\textwidth}
        \centering
        \includegraphics[width=1\textwidth]{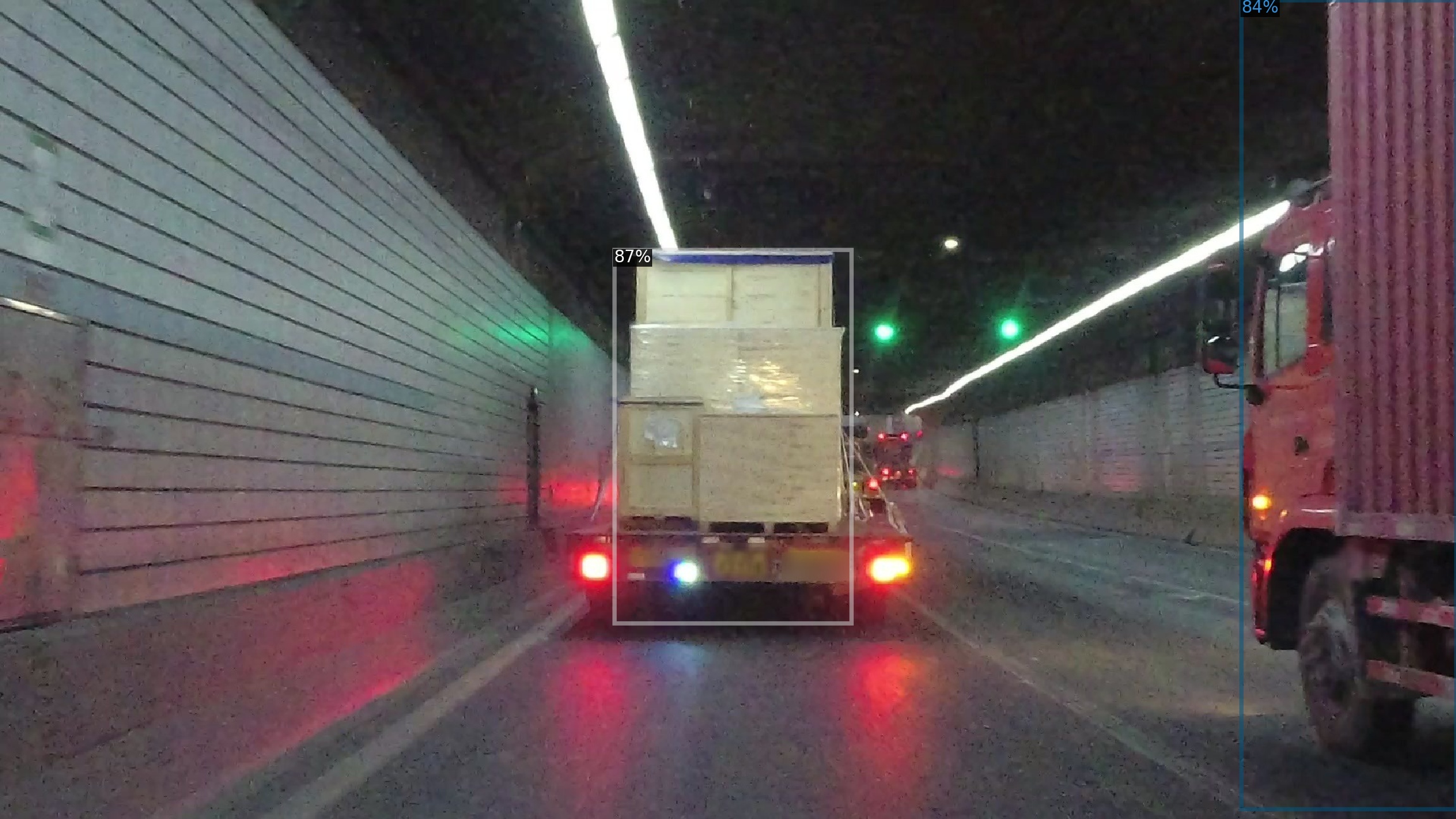}
        \includegraphics[width=1\textwidth]{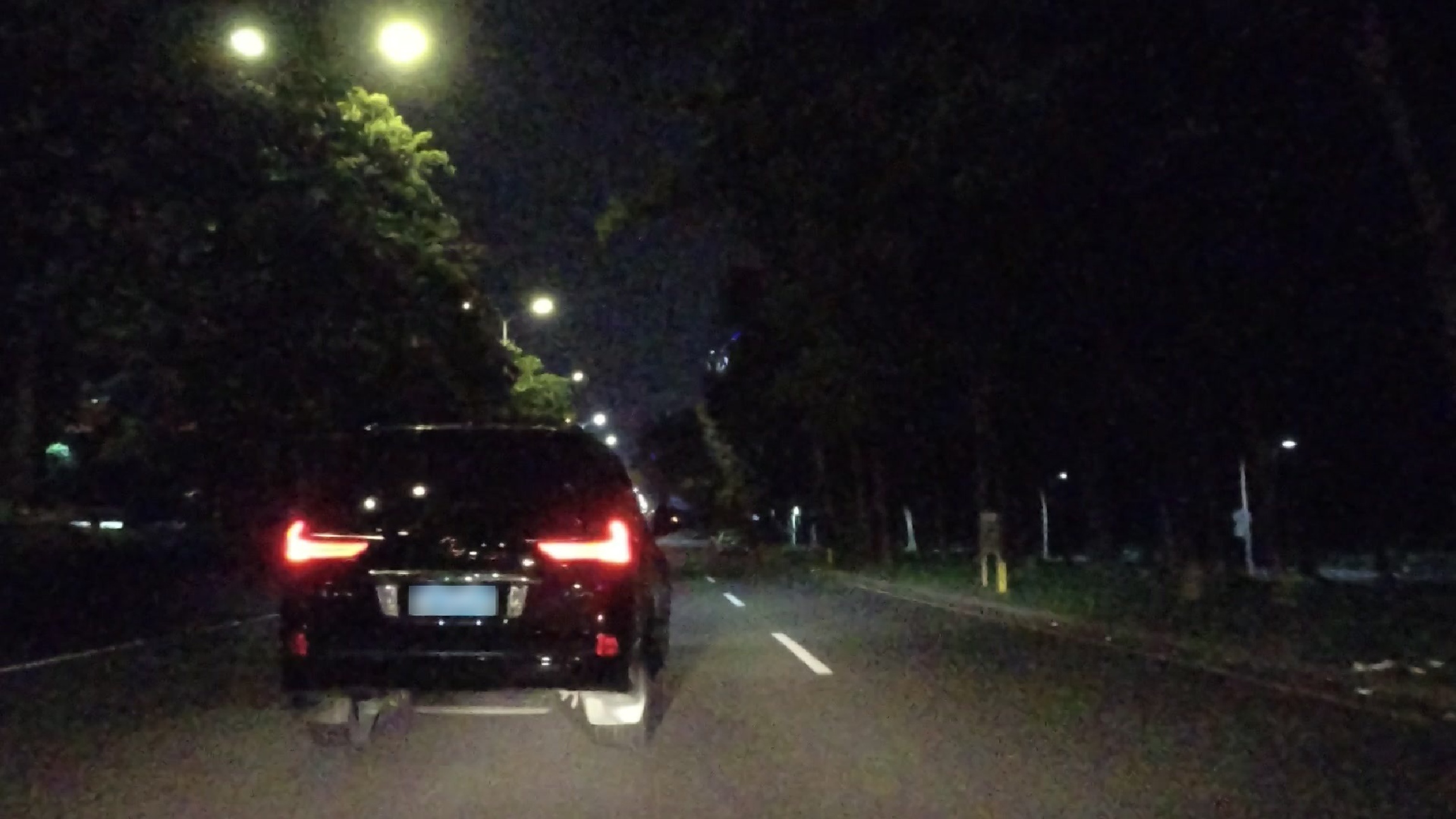}
        \caption{{\scriptsize ADePT (joint)}}
    \end{subfigure}
    \caption{Failure case under SODA10M~\cite{han2021soda10m}. Top: BYOL and SwAV exhibit truck detection failures. SoCo successfully detects three trucks. ADePT (sep) incorrectly predicts the small, distant truck as two vehicles while ADePT (joint) fails to detect. Bottom: BYOL and SwAV detect the car in the challenging night scenario, while SoCo and ADePT both fail. 
    \textbf{Best viewed with digital zoom}.}
    \label{fig:qualitative_soda:4}
\end{figure}

\subsubsection{Qualitative Examples}
In Fig.~\ref{fig:qualitative_soda} and Fig.~\ref{fig:qualitative_coco}, we present qualitative examples of model predictions on the SODA10M and MS COCO datasets. These examples confirm what is shown by our error analysis: ADePT tends to find most objects in an image including small objects instances that benefit from RPN pre-training.

In Fig.~\ref{fig:qualitative_coco:3} - Fig.~\ref{fig:qualitative_soda:4}, we provide additional visualizations to better understand the contributions and limitations of the proposed method. By comparing Fig.~\ref{fig:qualitative_coco:3} and Fig.~\ref{fig:qualitative_coco:4}, we conjecture that RPN pre-training helps more under normal imaging conditions when the pre-training and fine-tuning datasets have similar distribution. As shown in Fig.~\ref{fig:qualitative_soda:4}, ADePT fails under several challenging imaging conditions, such as darkness and overexposure, when the pre-training and fine-tuning datasets are drastically different. In practice, we suggest that the pre-training dataset should be selected to be similar to the fine-tuning dataset.

\section{Conclusion} 
\label{sec:conclusion}
In this work we present ADePT, a self-supervised learning strategy that pre-trains the regional proposal network (RPN) in a multi-stage detector. The alignment of both architecture and learning objectives result in improved downstream task performance. Through empirical study, we showed that (\textbf{A1}) RPN pre-training can reduce localization error, (\textbf{A2}) the alignment between the pretext and downstream tasks improve the overall performance, and (\textbf{A3}) RPN pre-training affords benefit to label-efficient scenarios. Future extension will improve the quality of unsupervised object proposals, beyond selective search, and leverage unsupervised mask labels towards instance segmentation training.

%
%
\bibliographystyle{elsarticle-num}
\bibliography{refs}

\begin{thebibliography}{10}
\expandafter\ifx\csname url\endcsname\relax
  \def\url#1{\texttt{#1}}\fi
\expandafter\ifx\csname urlprefix\endcsname\relax\def\urlprefix{URL }\fi
\expandafter\ifx\csname href\endcsname\relax
  \def\href#1#2{#2} \def\path#1{#1}\fi

\bibitem{he2020momentum}
K.~He, H.~Fan, Y.~Wu, S.~Xie, R.~Girshick, Momentum contrast for unsupervised
  visual representation learning, in: CVPR, 2020, pp. 9729--9738.

\bibitem{grill2020bootstrap}
J.-B. Grill, F.~Strub, F.~Altch{\'e}, C.~Tallec, P.~Richemond, E.~Buchatskaya,
  C.~Doersch, B.~Avila~Pires, Z.~Guo, M.~Gheshlaghi~Azar, et~al., Bootstrap
  your own latent-a new approach to self-supervised learning, in: NIPS,
  Vol.~33, 2020, pp. 21271--21284.

\bibitem{caron2021emerging}
M.~Caron, H.~Touvron, I.~Misra, H.~J{\'e}gou, J.~Mairal, P.~Bojanowski,
  A.~Joulin, Emerging properties in self-supervised vision transformers, in:
  ICCV, 2021, pp. 9650--9660.

\bibitem{ericsson2021well}
L.~Ericsson, H.~Gouk, T.~M. Hospedales, How well do self-supervised models
  transfer?, in: CVPR, 2021, pp. 5414--5423.

\bibitem{zhang2016colorful}
R.~Zhang, P.~Isola, A.~A. Efros, Colorful image colorization, in: ECCV,
  Springer, 2016, pp. 649--666.

\bibitem{dong2021self}
N.~Dong, M.~Kampffmeyer, I.~Voiculescu, Self-supervised multi-task
  representation learning for sequential medical images, in: ECML, Springer,
  2021, pp. 779--794.

\bibitem{wang2021dense}
X.~Wang, R.~Zhang, C.~Shen, T.~Kong, L.~Li, Dense contrastive learning for
  self-supervised visual pre-training, in: CVPR, 2021, pp. 3024--3033.

\bibitem{yang2021instance}
C.~Yang, Z.~Wu, B.~Zhou, S.~Lin, Instance localization for self-supervised
  detection pretraining, in: CVPR, 2021, pp. 3987--3996.

\bibitem{henaff2021efficient}
O.~J. H{\'e}naff, S.~Koppula, J.-B. Alayrac, A.~van~den Oord, O.~Vinyals,
  J.~Carreira, Efficient visual pretraining with contrastive detection, in:
  ICCV, 2021, pp. 10086--10096.

\bibitem{xie2021detco}
E.~Xie, J.~Ding, W.~Wang, X.~Zhan, H.~Xu, P.~Sun, Z.~Li, P.~Luo, Detco:
  Unsupervised contrastive learning for object detection, in: ICCV, 2021, pp.
  8392--8401.

\bibitem{xie2021propagate}
Z.~Xie, Y.~Lin, Z.~Zhang, Y.~Cao, S.~Lin, H.~Hu, Propagate yourself: Exploring
  pixel-level consistency for unsupervised visual representation learning, in:
  CVPR, 2021, pp. 16684--16693.

\bibitem{chen2021multisiam}
K.~Chen, L.~Hong, H.~Xu, Z.~Li, D.-Y. Yeung, Multisiam: Self-supervised
  multi-instance siamese representation learning for autonomous driving, in:
  ICCV, 2021, pp. 7546--7554.

\bibitem{wei2021aligning}
F.~Wei, Y.~Gao, Z.~Wu, H.~Hu, S.~Lin, Aligning pretraining for detection via
  object-level contrastive learning, in: NIPS, Vol.~34, 2021, pp. 22682--22694.

\bibitem{caron2020unsupervised}
M.~Caron, I.~Misra, J.~Mairal, P.~Goyal, P.~Bojanowski, A.~Joulin, Unsupervised
  learning of visual features by contrasting cluster assignments, in: NIPS,
  Vol.~33, 2020, pp. 9912--9924.

\bibitem{han2021soda10m}
J.~Han, X.~Liang, H.~Xu, K.~Chen, H.~Lanqing, J.~Mao, C.~Ye, W.~Zhang, Z.~Li,
  X.~Liang, et~al., Soda10m: A large-scale 2d self/semi-supervised object
  detection dataset for autonomous driving, in: NIPS Track on Datasets and
  Benchmarks, 2021.

\bibitem{bolya2020tide}
D.~Bolya, S.~Foley, J.~Hays, J.~Hoffman, {TIDE}: A general toolbox for
  identifying object detection errors, in: ECCV, 2020, pp. 558--573.

\bibitem{deng2009imagenet}
J.~Deng, W.~Dong, R.~Socher, L.-J. Li, K.~Li, L.~Fei-Fei, Imagenet: A
  large-scale hierarchical image database, in: CVPR, IEEE, 2009, pp. 248--255.

\bibitem{he2017mask}
K.~He, G.~Gkioxari, P.~Doll{\'a}r, R.~Girshick, Mask r-cnn, in: ICCV, 2017, pp.
  2961--2969.

\bibitem{ren2015faster}
S.~Ren, K.~He, R.~Girshick, J.~Sun, Faster r-cnn: Towards real-time object
  detection with region proposal networks, in: NIPS, Vol.~28, 2015, pp. 91--99.

\bibitem{uijlings2013selective}
J.~R. Uijlings, K.~E. Van De~Sande, T.~Gevers, A.~W. Smeulders, Selective
  search for object recognition, IJCV 104~(2) (2013) 154--171.

\bibitem{lin2014microsoft}
T.-Y. Lin, M.~Maire, S.~Belongie, J.~Hays, P.~Perona, D.~Ramanan,
  P.~Doll{\'a}r, C.~L. Zitnick, Microsoft coco: Common objects in context, in:
  ECCV, Springer, 2014, pp. 740--755.

\bibitem{everingham2010pascal}
M.~Everingham, L.~Van~Gool, C.~K. Williams, J.~Winn, A.~Zisserman, The pascal
  visual object classes (voc) challenge, IJCV 88~(2) (2010) 303--338.

\bibitem{girshick2014rich}
R.~Girshick, J.~Donahue, T.~Darrell, J.~Malik, Rich feature hierarchies for
  accurate object detection and semantic segmentation, in: CVPR, 2014, pp.
  580--587.

\bibitem{girshick2015fast}
R.~Girshick, Fast r-cnn, in: ICCV, 2015, pp. 1440--1448.

\bibitem{cai2018cascade}
Z.~Cai, N.~Vasconcelos, Cascade r-cnn: Delving into high quality object
  detection, in: CVPR, 2018, pp. 6154--6162.

\bibitem{henaff2020data}
O.~Henaff, Data-efficient image recognition with contrastive predictive coding,
  in: ICML, PMLR, 2020, pp. 4182--4192.

\bibitem{chen2020simple}
T.~Chen, S.~Kornblith, M.~Norouzi, G.~Hinton, A simple framework for
  contrastive learning of visual representations, in: ICML, PMLR, 2020, pp.
  1597--1607.

\bibitem{lin2017feature}
T.-Y. Lin, P.~Doll{\'a}r, R.~Girshick, K.~He, B.~Hariharan, S.~Belongie,
  Feature pyramid networks for object detection, in: CVPR, 2017, pp.
  2117--2125.

\bibitem{bai2022point}
Y.~Bai, X.~Chen, A.~Kirillov, A.~Yuille, A.~C. Berg, Point-level region
  contrast for object detection pre-training, in: CVPR, 2022, pp. 16061--16070.

\bibitem{islam2023self}
A.~Islam, B.~Lundell, H.~Sawhney, S.~N. Sinha, P.~Morales, R.~J. Radke,
  Self-supervised learning with local contrastive loss for detection and
  semantic segmentation, in: WACV, 2023, pp. 5624--5633.

\bibitem{he2016deep}
K.~He, X.~Zhang, S.~Ren, J.~Sun, Deep residual learning for image recognition,
  in: CVPR, 2016, pp. 770--778.

\bibitem{pont2015boosting}
J.~Pont-Tuset, L.~Van~Gool, Boosting object proposals: From pascal to coco, in:
  ICCV, 2015, pp. 1546--1554.

\bibitem{wang2020frustratingly}
X.~Wang, T.~Huang, J.~Gonzalez, T.~Darrell, F.~Yu, Frustratingly simple
  few-shot object detection, in: ICML, PMLR, 2020, pp. 9919--9928.

\bibitem{fan2021generalized}
Z.~Fan, Y.~Ma, Z.~Li, J.~Sun, Generalized few-shot object detection without
  forgetting, in: CVPR, 2021, pp. 4527--4536.

\bibitem{vinyals2016matching}
O.~Vinyals, C.~Blundell, T.~Lillicrap, K.~Kavukcuoglu, D.~Wierstra, Matching
  networks for one shot learning, in: Advances in Neural Information Processing
  Systems, 2016, pp. 3637--3645.

\bibitem{snell2017prototypical}
J.~Snell, K.~Swersky, R.~Zemel, Prototypical networks for few-shot learning,
  in: Advances in Neural Information Processing Systems, Vol.~34, 2017, pp.
  4080--4090.

\bibitem{dong2022residual}
N.~Dong, M.~Maggioni, Y.~Yang, E.~P{\'e}rez-Pellitero, A.~Leonardis,
  S.~McDonagh, Residual contrastive learning for image reconstruction: Learning
  transferable representations from noisy images, in: IJCAI, 2022, pp.
  2930--2936.

\end{thebibliography}

\end{document}